\documentclass[1x0pt,twocolumn,letterpaper]{article}

\usepackage[algorithms]{wacv}  %

\usepackage{breqn}
\usepackage{amsmath,amsfonts,bm}
\usepackage{comment}
\usepackage{stmaryrd}
\usepackage{subcaption}
\usepackage{graphicx}
\usepackage[T1]{fontenc}
\usepackage[htt]{hyphenat}
\usepackage[nowatermark]{fixmetodonotes} %

\usepackage[dvipsnames]{xcolor}
\usepackage{colortbl}
\usepackage{pifont}
\usepackage{multirow}
\usepackage{caption} 
\usepackage{tcolorbox}

\newcommand{\method}{VIG-TUQ}

\usepackage{soul}

\def\vtheta{{\bm{\theta}}}

\def\valpha{{\bm{\alpha}}}

\def\eqref#1{equation~\ref{#1}}

\def\1{\bm{1}}

\def\vtheta{{\bm{\theta}}}

\def\vh{{\bm{h}}}

\def\vq{{\bm{q}}}

\def\vs{{\bm{s}}}

\def\vx{{\bm{x}}}
\def\vy{{\bm{y}}}

\def\mK{{\bm{K}}}

\DeclareMathAlphabet{\mathsfit}{\encodingdefault}{\sfdefault}{m}{sl}
\SetMathAlphabet{\mathsfit}{bold}{\encodingdefault}{\sfdefault}{bx}{n}

\newcommand{\R}{\mathbb{R}}

\definecolor{wacvblue}{rgb}{0.21,0.49,0.74}
\usepackage[pagebackref,breaklinks,colorlinks,allcolors=wacvblue]{hyperref}
\newcommand{\joseph}[1]{\textcolor{black}{#1}}
\def\wacvPaperID{611} %
\def\confName{WACV}
\def\confYear{2027}

\title{Leveraging Visual Signals for Robust Token-Level Uncertainty in Vision–Language Generation}

\author{
Joseph Hoche\textsuperscript{1} \quad
David Brellmann\textsuperscript{2} \quad
Gianni Franchi\textsuperscript{1,3}\\[0.3em]
\textsuperscript{1}AMIAD, Pôle Recherche, Palaiseau \quad
\textsuperscript{2}Safran Tech \quad
\textsuperscript{3}ENSTA Paris\\
}

\begin{document}
\maketitle
\begin{abstract}
  Uncertainty Quantification (UQ) remains a critical challenge in Large Vision Language Models (LVLMs) for reliable predictions and real-world deployment. 
    However, most existing methods are adapted from the LLM literature and primarily focus on the language modality, leaving the contribution of visual information to LVLM uncertainty largely underexplored.
    In this paper, we investigate how LVLMs process visual information and whether this process can be used to improve uncertainty estimation.
    By analyzing hidden representations after the integration of visual features during the generation process, we observe that high-confidence predictions rely more heavily on visual content than uncertain ones.
    Building on this insight, we propose \textbf{Visual-Grounded Token UQ (VIG-TUQ)}, a training-free framework that explicitly incorporates visual grounding into uncertainty estimation by weighting token-level language uncertainty with visual grounding scores.
    We evaluate VIG-TUQ on multiple datasets and across diverse LVLM architectures, including early-fusion, late-fusion, and native-fusion models. 
    Results indicate that our method often improves upon existing token-level uncertainty approaches.
Code and data will be made available upon acceptance.
    
\end{abstract}

\section{Introduction}
\label{Introduction}

Large Vision-Language Models (LVLMs) extend Large Language Models (LLMs) with visual perception and excel across a wide spectrum of multimodal tasks requiring visual and language understanding~\citep{openai2024gpt4, geminiteam2024, guo2025deepseek, openai2025gptoss, novikov2025alphaevolve}. 
Despite their impressive abilities, LVLMs are not yet reliable.
In particular, they remain prone to hallucinations, i.e., they may generate incorrect, incomplete, fabricated, or misleading content that appears plausible~\citep{ji2023, liu2024survey, huang2025}.
Such behavior limits their use in safety-critical applications like autonomous driving~\citep{gao2025survey}, robotics~\citep{han2025multimodal} or medicine~\citep{jiang2024evaluating}.
A promising direction for improving deployment reliability is uncertainty quantification (UQ) to detect when generated content may be unreliable~\citep{hernandez2015, gal2016dropout, abdar2021}.
Prior works show promising results for UQ methods in the language setting for LLMs~\citep{farquhar2024detecting, nikitin2024kernel, chen2024}.
Although these approaches can be extended to LVLMs~\citep{zhang2024vluncertainty, lau2025uncertainty}, their extension is not straightforward.
LVLMs operate across multiple modalities and their uncertainty may stem from both linguistic and visual sources.
Recent studies have highlighted the tendency of LVLMs to prioritize internal textual knowledge over external visual information~\citep{leeetal2025, luo2025probing, fu2025, long2026understanding}.
For instance, given an image depicting a blue orange and the question \textit{``what color is the orange?''}, an LVLM may confidently answer \textit{``orange''}, basing its answer on its parametric knowledge and language prior rather than on the actual visual input. 
This mismatch motivates the importance of developing specific approaches designed for LVLMs that explicitly capture and quantify the contribution of visual information to LVLM predictions.

\begin{figure*}[t]
    \centering
    \includegraphics[width=0.7\textwidth]{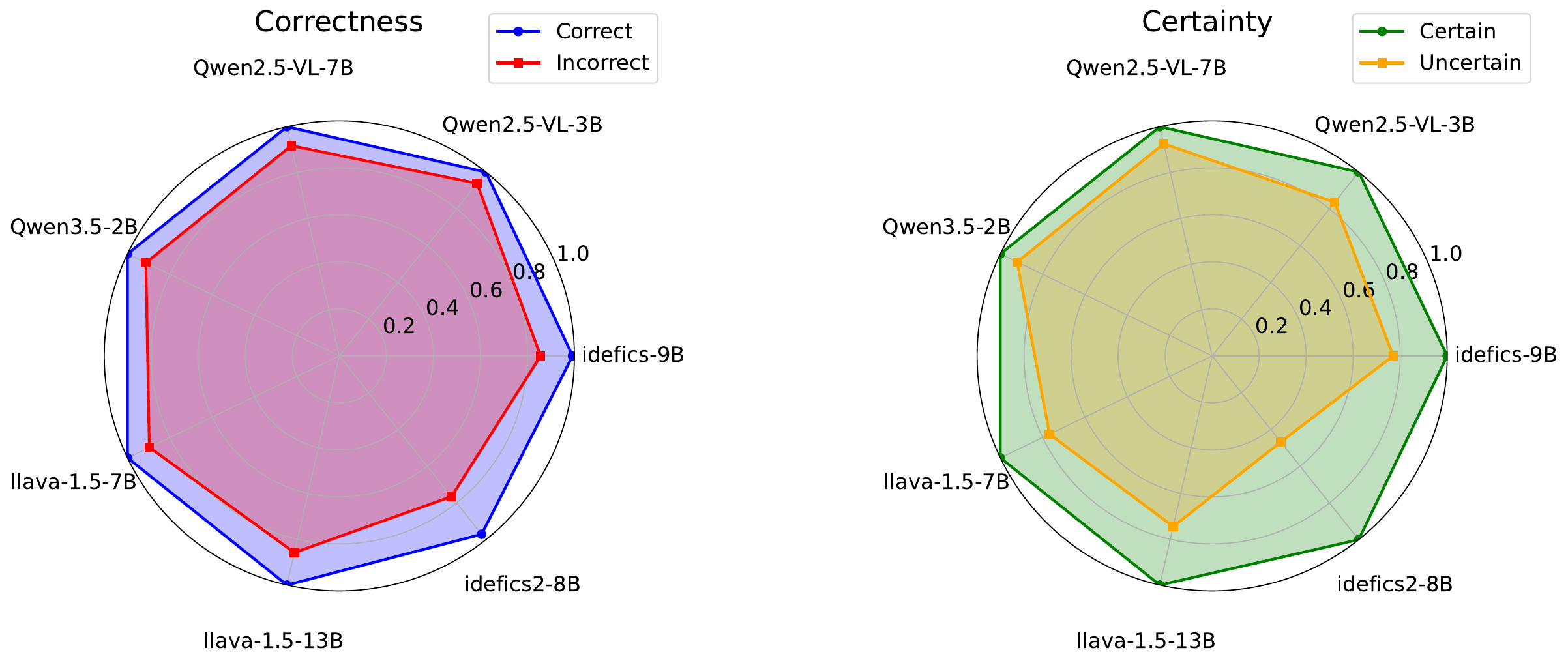}
    \caption{
    \textbf{Correct / confident predictions depend more strongly on visual information than incorrect / uncertain ones.}
    Radial values indicate the cosine distance between hidden representations from two forward passes: one with visual input and one without. 
    Results are averaged on the \textbf{OKVQA} dataset~\citep{marino2019ok}.
    The cosine distance measures the model reliance on visual information, where larger values indicate a stronger influence of the image on the internal representation.
    }
      \vspace{-15pt}
    \label{fig:spider}

\end{figure*}

\begin{figure*}[t!]
    \centering
    \vspace{-15pt}
    \begin{center}
        \includegraphics[width=0.8\linewidth]{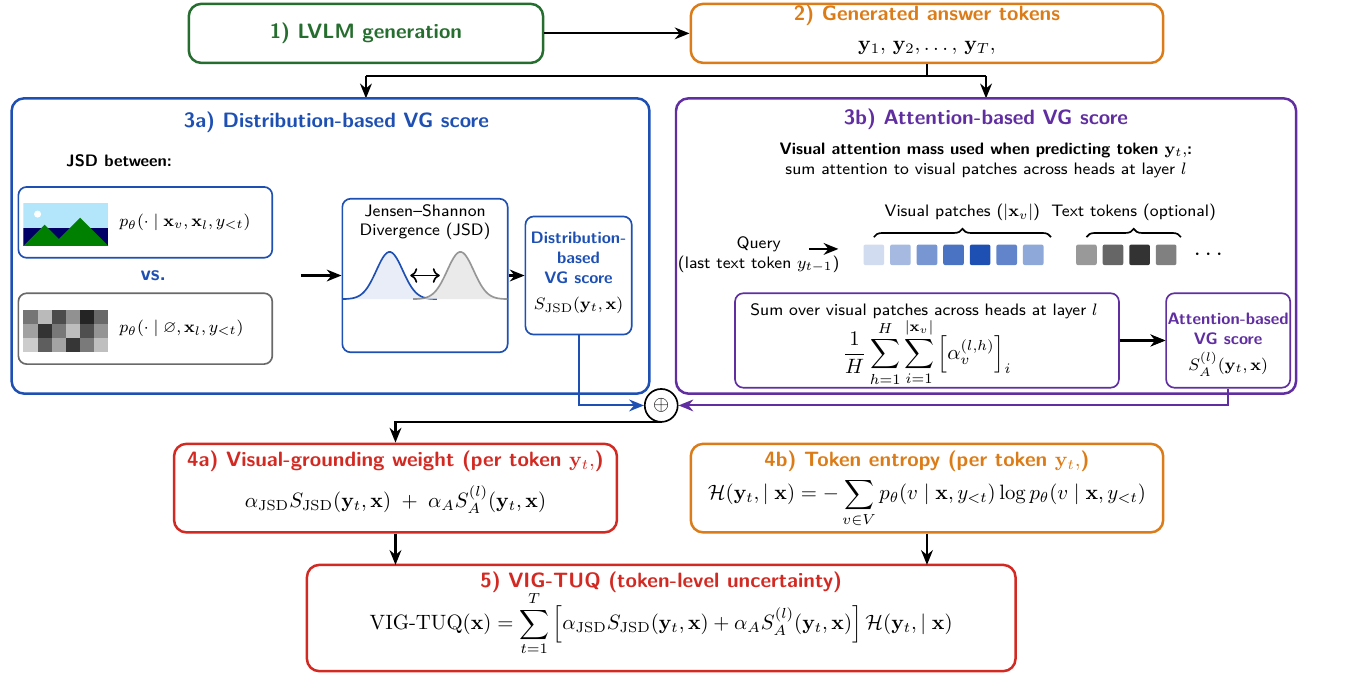}
    \end{center}
    \vspace{-15pt}
    \caption{\textbf{Overview of the VIG-TUQ pipeline.} 
    VIG-TUQ weights token-level language uncertainty with visual grounding scores using two complementary strategies: a distribution-based score obtained from the Jensen–Shannon divergence between predictions with and without the image, and an attention-based score measuring the attention mass assigned to visual patches.}
    \vspace{-15pt}
    \label{fig:vision_attention}
\end{figure*}

In this paper, we investigate how LVLM process visual information during next-token prediction and whether this process can be leveraged to improve token-level uncertainty estimation. 
By analyzing hidden representations during generation, we observe that high-confidence predictions rely more heavily on visual content than uncertain ones (Figure~\ref{fig:spider}).
Building on this insight, we propose \textbf{VIsual-Grounding Token Uncertainty Quantification (VIG-TUQ)}, a training-free uncertainty quantification framework for LVLMs. 
VIG-TUQ explicitly incorporates visual grounding into uncertainty estimation by weighting token-level language uncertainty with visual grounding scores (Figure~\ref{fig:vision_attention}).
Specifically, VIG-TUQ quantifies the influence of visual content on next-token prediction using two complementary visual grounding strategies: 
(1) comparing next-token distributions with and without visual input, and 
(2) leveraging visual attention weights over image tokens. 
\joseph{Our central hypothesis is that token-level uncertainties should not necessarily contribute equally to the uncertainty of the full generated sequence. Visual grounding is therefore used as a token-importance signal: it determines how strongly each token-level uncertainty contributes to the final sequence-level score, rather than acting as an uncertainty measure by itself.} 
Our experimental results support this hypothesis and highlight the importance of combining multiple visual grounding signals, as their effectiveness can be model-dependent and capture distinct aspects of visual reasoning.
Finally, our approach provides a fine-grained into which specific spans or entities contribute to the uncertainty, while remaining computationally efficient by avoiding reliance on external models or sampling.

\vspace{-10pt}
\paragraph{Contributions.}

This work makes the following key contributions:
(1) We observe that LVLMs tend to be more confident when their predictions rely more strongly on visual input. 
(2) We introduce a new training-free uncertainty quantification framework that combines different complementary visual grounding scores to weight token-level uncertainty measures.
We evaluate VIG-TUQ on multiple datasets and across diverse LVLM architectures, including early-fusion, late-fusion, and native-fusion models. 
Results show that our method often improves upon existing token-level uncertainty approaches and remain computationally efficient due to the absence of external models or sampling.
(3) We analyze whether uncertainty is uniformly distributed across generated tokens or concentrated within a subset of informative tokens. 
Our findings indicate that visual information helps identify the tokens most relevant for uncertainty estimation, enabling more effective token selection in LVLMs for uncertainty estimation.

\section{Related work}
\label{Related_Work}

Several recent methods have been proposed to enhance reliability by estimating uncertainty in the predictions of LLMs and LVLMs. 
These methods are broadly categorized as follows~\citep{kang2025uncertainty}:
(1) logit-based methods, which directly use token probabilities/logits as uncertainty scores~\citep{malinin2021uncertainty, Fomicheva2020}; 
(2) semantic-based methods, which quantifies uncertainty by assessing the semantic consistency of multiple generated answers~\citep{kuhn2023semantic, farquhar2024detecting, zhang2024vluncertainty, lin2024generating, grewal2024improving, lau2025uncertainty, joocho2025};
(3) verbalized-based methods, which estimate uncertainty by directly querying the LLM about the reliability of its own generations~\citep{kadavath2022language, yona2024can, xiong2024can};
and (4) internal-state-based methods, which analyze internal signals from LLMs after a single generation to estimate uncertainty, such as features at intermediate layers, attention weights, or other latent representations~\citep{azaria2023the, orgad2025llms, kim2025}.
Methods listed above may rely on external models (e.g., sentence embedding models~\citep{grewal2024improving, abdaljalil2025}, NLI models~\citep{fadeeva2024fact, nikitin2024kernel}, instruction-tuned LLMs~\citep{farquhar2024detecting, ji2025calibrating}) or involve training dedicated models to estimate uncertainty~\citep{fieback2024metatoken, orgad2025llms, hoche2025improving}.
In this paper, we propose a training-free hybrid approach to uncertainty estimation that combines a logit-based component to capture linguistic uncertainty with an internal-state-based component to incorporate visual uncertainty. 
By explicitly leveraging visual information while maintaining computational efficiency with a single-pass and token-level estimation, we introduce a multimodal-aware approach designed for LVLMs. 
Additional technical details and relevant results from prior work are discussed throughout the paper.

\section{Preliminaries}
\label{sec:preliminaries}
\paragraph{Notations.}
Let $\mathcal{X}_v$ denote the visual input space, $\mathcal{X}_l$ the linguistic input space, and $\mathcal{V}$ the LVLM vocabulary.
A multimodal input is depicted as $\vx=(\vx_v, \vx_l)$, where the visual component $\vx_v \in \mathcal{X}_v$ consists of a sequence of visual patches and the textual component $\vx_l \in \mathcal{X}_l$ consists of a sequence of textual tokens.
\newline
\newline
An LVLM defines a parameterized function $f_\vtheta: \mathcal{X}_v \times \mathcal{X}_l \to \R^d$ to encode the visual-linguistic context into a latent representation.
In practice, $f_\vtheta(\cdot, \cdot)$ typically includes a transformer decoder architecture, potentially combined with separate encoders for each modality (see Section~\ref{sec:modality_fusion}). 
For a given multimodal input $\vx=(\vx_v, \vx_l)$, the model then autoregressively generates a token sequence $\vy=\bigl[\vy_t\bigr]_{t=1}^T$, where each token $\vy_t$ is sampled according to:
\begin{equation}
    p_\vtheta(\vy_t \mid \vx, \vy_{<t})=\operatorname{softmax}\Bigl(f_h \bigl(f_\vtheta\bigl(\vx_v, [\vx_l \mid \vy_{<t}]\bigr)\bigr)\Bigr),
    \label{def:lvlm_distrib}
\end{equation}
where  $f_h: \R^d \to \R^{\lvert \mathcal{V} \rvert}$ denotes the output head that projects the latent representation to logits over the vocabulary $\mathcal{V}$ and $\vy_{<t}=\bigl[\vy_1, \ldots, \vy_{t-1}\bigr]$ depicts the previously generated tokens.
\subsection{Modality fusion strategies}
\label{sec:modality_fusion}
Given an input $\vx=(\vx_v, \vx_l)$, an LVLM can be described by the approach it uses to combine information from the visual modality $\vx_v \in \mathcal{X}_v$ and the textual modality $\vx_{l} \in \mathcal{X}_l$ within the parameterized function $f_\vtheta: \mathcal{X}_v \times \mathcal{X}_l \to \R^d$.
\paragraph{Native fusion.}
In native fusion, the modalities are directly combined into a single representation before being processed by the transformer decoder architecture. 
In particular, the parametric function $f_\vtheta: \mathcal{X}_v \times \mathcal{X}_l \to \R^d$ is expressed as a composition  $f_\vtheta=f_{L} \circ f_{L-1} \circ \ldots \circ f_{1} \circ f_{0}$, where $f_{0}: \mathcal{X}_v \times \mathcal{X}_l \to \R^d$ is  the input function and $\{f_{i}: \R^d \to \R^d\}_{i=1}^L$ are $L$ decoder transformer layers.
The input function $f_{0}(\cdot, \cdot)$ processes image patches and text tokens independently as:
\begin{equation}
    f_{0}(\vx_v,\vx_l) = \bigl[E_v(\vx_{v}) \mid E_l(\vx_{l})\bigr],
\end{equation}
where $E_v(\cdot)$ and $E_l(\cdot)$ map visual patches and language tokens to their embeddings via linear transformations.
In these architectures, the model is trained end-to-end from scratch to learn multimodal representations natively within the transformer blocks.
LVLM like Qwen models~\citep{qwen3.5} adopts a native fusion strategy.
\paragraph{Early fusion.}
In early fusion, features from each modality are first extracted separately using dedicated encoders, and then integrated into a shared representation before being processed by the transformer decoder architecture. 
In particular, the parametric function $f_\vtheta: \mathcal{X}_v \times \mathcal{X}_l \to \R^d$ is defined as a composition  $f_\vtheta=f_L \circ f_{L-1} \circ \ldots \circ f_{1} \circ f_{0}$, where $\{f_{i}: \R^d \to \R^d\}_{i=1}^L$ are $L$ decoder transformer layers and $f_{0}: \mathcal{X}_v \times \mathcal{X}_l \to \R^d$ is the input function defined as:
\begin{equation}
    f_{0}(\vx_v,\vx_l) = m_{0}\bigl(E_{\vtheta_v}(\vx_v), E_l(\vx_l)\bigr);
\end{equation}
with $m_{0}(\cdot, \cdot)$ a multimodal fusion module (e.g., concatenation, linear layer, cross-attention layer);
$E_{\vtheta_v}(\cdot)$ a visual encoder with parameters $\vtheta_v$ (e.g., ViT, CNNs);
and $E_l(\cdot)$ a text embedding function.
Unlike native fusion, the visual encoder $E_{\vtheta_v}(\cdot)$ is a pre-trained model that can be optimized independently.
LVLMs like LLaVA~\citep{liu2023visual} and Qwen-VL~\citep{bai2024qwenvl} adopt an early fusion approach.
Further details on early fusion, fusion modules, and unimodal encoders can be found in~\citep{li2024multimodal, an2025towards}.

\paragraph{Late fusion.}
Late fusion also starts with the extraction of unimodal features using dedicated encoders. 
However, unlike early fusion, the integration between modalities occurs within the intermediate representations of the transformer decoder architecture rather than before them.
In particular, the parametric function $f_\vtheta: \mathcal{X}_v \times \mathcal{X}_l \to \R^d$ can be defined recursively over $L$ layers as
\begin{align}
     \begin{split}
         f_{\vtheta}(\vx_v,\vx_l) &= f_{L}(\vh_{L-1},\vx_v), \qquad \text{where} \\
    \vh_i &= f_{i}(\vh_{i-1},\vx_v) = g_i\bigl(m_{i}\bigl(E_{\vtheta_v^{(i)}}(\vx_v), \vh_{i-1}\bigr)\bigr)  \\  (1 \leq i \leq L)  \vh_0 &=  E_l(\vx_{l});
     \end{split}
\end{align}
with  $E_l(\cdot)$ a text embedding function. 
For each layer $i$, $g_i: \R^d \to \R^d$ depicts a transformer decoder layer, $m_{i}(\cdot, \cdot)$ denotes a multimodal fusion module (e.g., concatenation, linear layer, cross-attention layer), and $E_{\vtheta_v^{(i)}}(\cdot)$ is a visual encoder with parameters $\vtheta_v^{(i)}$ (e.g., ViT, CNNs).
Note that the vision modality is not necessarily integrated at every layer and the fusion module can simply preserve the hidden representation, i.e., $m_{i}\bigl(E_{\vtheta_v^{(i)}}(\vx_v), \vh_i\bigr)=\vh_i$.
Similarly to early fusion, visual encoders $E_{\vtheta_v^{(i)}}(\cdot)$ are pre-trained models that can be optimized independently.
LVLMs like IDEFICS~\citep{laurenccon2023obelics}
and Flamingo~\citep{alayrac2022flamingo} adopt such approach.
Further details on late fusion, fusion modules, and unimodal encoders can be found in~\citep{li2024multimodal, an2025towards}.
\subsection{Token-level UQ methods}
\label{sec:token_level_uq}
Given a multimodal input $\vx$, token-level UQ methods assign an uncertainty score $\vs_t$ to each generated token $\vy_t$ (\eqref{def:lvlm_distrib}), with a higher score indicating greater uncertainty. 
The uncertainty score $s(\cdot)$ for a generated sequence $\vy=[\vy_t]_{t=1}^T$ can be derived by aggregating the token-level scores of each token $\vy_t$, e.g., by averaging across tokens $s(\vy) = \tfrac{1}{T} \sum_{t=1}^{T} \vs_t$ or by taking their maximum $s(\vy) = \max_t \vs_t$.
\paragraph{Negative Log-Likelihood (Log-Perplexity).} 
A straightforward token-level uncertainty score is the token negative log-likelihood~\citep{Fomicheva2020, malinin2021}:
\begin{equation}
    \operatorname{NLL}(\vy_t, \vx, \vy_{<t})= - \log p_\vtheta(\vy_t \mid \vx, \vy_{<t}).
     \label{def:LP}
\end{equation}
\paragraph{Token Entropy.}
Another common token-level score is the token entropy of the autoregressive distribution~\citep{Fomicheva2020, malinin2021}:
\begin{equation}
    \mathcal{H}(\vy \mid \vx, \vy_{<t}) =  - \sum_{\vy \in \mathcal{V}} p_\vtheta(\vy \mid \vx, \vy_{<t}) \log p_\vtheta(\vy \mid \vx, \vy_{<t}).   
    \label{def:token_entropy}
\end{equation}
\paragraph{Claim-Conditioned Probability (CCP).}
CCP~\citep{fadeeva2024fact} estimates token-level uncertainty by measuring how much the top $K$ token alternatives $\{\vy_t^k\}_{k=1}^K$ contradict the generated token $\vy_t$ using a Natural Language Inference (NLI) model:

\begin{multline}
    \operatorname{CCP}(\vy_t, \vx, \vy_{<t}) = \\
    - \log  \frac{\sum_{k: \text{NLI}(\vy_{<t}, \vy_t^k, \vy_t)=\text{entail}}  p_\vtheta(\vy_t^k \mid \vx, \vy_{<t})}{\sum_{k: \text{NLI}(\vy_{<t}, \vy_t^k, \vy_t)\in\{\text{entail},\text{contradict}\}}  p_\vtheta(\vy_t^k \mid \vx, \vy_{<t})},
    \label{def:CCP}
\end{multline}
where $\text{NLI}(\cdot)$ determines whether concatenating $\vy_t^k$ with the preceding generated context $\vy_{<t}$ entails (semantic equivalence) or contradicts (different meaning) the generated content $\vy_{<t} \cup \vy_t$. 

\paragraph{Token-Shifting Attention to more Relevant (Token-SAR).}
Token-SAR~\citep{duan2024shifting} defines a token-level uncertainty score by reweighting token likelihood with a semantic relevance factor as:
\begin{equation}
    \operatorname{Token-SAR}(\vy_t, \vx, \vy_{<t})= - \log p_\vtheta(\vy_t \mid \vx, \vy_{<t}) \, \tilde{R}(\vx, \vy, \vy_t),
    \label{def:SAR}
\end{equation}
where $\tilde{R}(\vx, \vy, \vy_t)=R(\vx, \vy, \vy_t)/\sum_{i=1}^TR(\vx, \vy, \vy_i)$ is the normalized relevance score with 
$$
    R(\vx, \vy, \vy_t)=1-\bigl\lvert g(\vx \cup \vy, \vx \cup \vy \backslash \{\vy_t\}) \bigr\rvert
$$ 
the relevance that quantifies how important $\vy_t$ is in reflecting the semantics of $\vy$ by comparing the semantic change before and after removing this token and  $g(\cdot, \cdot)$ a similarity between two inputs on a scale of $-1$ to $1$.

\section{Method}
\label{sec:methods}

In this section, we introduce VIsual-Grounding Token Uncertainty Quantification (\method), a training-free UQ framework for LVLMs that weights token-level language uncertainty with visual grounding scores. 
We first discuss our motivations (Section~\ref{sec:motivations}), then describe the visual grounding score used (Section~\ref{sec:vgs}), and finally introduce the proposed method (Section~\ref{sec:method}).

\subsection{Motivations}
\label{sec:motivations}

A recent line of works highlights the tendency of LVLMs to prioritize internal textual knowledge over external visual information~\citep{leeetal2025, luo2025probing, fu2025, long2026understanding}.
This suggests that visual grounding could play a crucial role in identifying unreliable responses.
To investigate this phenomenon and better understand the contribution of visual information to token generations, we conduct a pilot study on how LVLMs process visual information.
Specifically, we compare hidden representations (via the cosine distance) obtained from two independent forward passes: one with visual evidence and one without it.
This analysis is performed on generated sentences from the \textbf{OKVQA} dataset~\citep{marino2019ok} across different LVLM architectures~\citep{long2026understanding} using greedy decoding with visual evidence 
(further details are provided in Appendix~\ref{sec:architecture}).
Figure~\ref{fig:spider} summarizes our results and highlights that correct and confident predictions rely more on the visual content than incorrect and uncertain ones.
Furthermore, experiments reported in Appendix~\ref{sec:app_Additional_Experiments} and Appendix~\ref{sec:architecture} indicate that the extent to which visual information is processed depends on both the training setup and the model architecture (see Section~\ref{sec:preliminaries}). 
These observations motivate the introduction of multiple visual grounding scores to capture different aspects of visual understanding. 
Overall, our results suggest that tokens whose generation strongly depends on visual input are more informative and should therefore be assigned greater weight in uncertainty estimation.
Building on this insight, we introduce diffrent visual grounding token-level strategies and our VIG-TUQ approach in the following sections.

\subsection{Visual grounding scores}
\label{sec:vgs}
We propose two scoring strategies to measure how much an LVLM relies on its visual input when generating a token.

\paragraph{Distribution-based scores.}
Given a multimodal input $\vx=(\vx_v,\vx_l)$, with visual content $\vx_v$ and textual content $\vx_l$, this approach quantifies the contribution of the visual input $\vx_v$ by measuring how much the next-token distribution changes when the visual information is removed. 
In particular, for each generated token $\vy_t$, we compare the original LVLM next-token distribution
$p_\vtheta(\cdot \mid \vx_v,\vx_l,\vy_{<t})$
with the next-token distribution $p_\vtheta(\cdot \mid \varnothing,\vx_l,\vy_{<t})$ when the visual information $\vx_v$ is removed using the Jensen-Shannon divergence:
\begin{equation}
    d_t(\vy,\vx)
    =
    \operatorname{JSD}
    \left(
        p_\vtheta(\cdot \mid \vx_v,\vx_l,\vy_{<t})
        \,\middle\|\,
        p_\vtheta(\cdot \mid \varnothing,\vx_l,\vy_{<t})
    \right).
    \label{def:js_raw_score}
\end{equation}

A larger value of $d_t(\vy,\vx)$ indicates that removing the visual input induces a stronger distributional shift when generating $\vy_t$ and thus a greater reliance on visual information. 
To obtain a relative token-level importance score, we normalize the divergence values across all generated tokens: $S_{\mathrm{JSD}}(y_t,\vx)$, where 
\begin{equation}
    S_{\mathrm{JSD}}(\vy_t,\vx)=\frac{d_t(\vy_t,\vx)}{\sum_{i=1}^Td_t(\vy_i,\vx)}.
    \label{def:js_raw_score}
\end{equation}
This normalized score captures the relative contribution of the visual input to each generated token $\vy_t$ with respect to the full sequence.

\paragraph{Attention-based Scores.}
The attention mechanism provides another way to estimate how much the visual content contributes to the next-token prediction. 
For instance, given a multimodal input $\vx=(\vx_v, \vx_l)$ with a visual content $\vx_v$ and a textual content $\vx_l$, we could use the attention weight vector used for predicting the next token in the $l$-th layer and the $h$-th attention head that interacts with the visual content $\vx_v$:
\begin{equation}
    \valpha_t^{(l,h)}
    =
    \operatorname{softmax}
    \Bigl(
        \tfrac{\mK_t^{(l,h)}\vq_t^{(l,h)}}{\sqrt{d_k}}
    \Bigr)
    \in [0,1]^N,
    \label{def:attention_weight_vector}
\end{equation}
where $\vq^{(l,h)} \in \R^{d_k}$ is the query vector corresponding to the last textual token in $\vx_l$;  
$\mK^{(l,h)} \in \R^{N \times d_k}$ is the key matrix featuring the visual content $\vx_v$ and possibly the textual content $\vx_l$ (in the case of self-attention); 
$d_k$ is the dimensionality of the key and query vectors; 
and $N$ is the number of visual (and possibly textual) elements in $\vx$.
We denote by $\valpha_{t,v}^{(l,h)}$ the subset of attention weights interacting with visual tokens or patches. 
For cross-attention heads, $\valpha_{t,v}^{(l,h)}=\valpha_t^{(l,h)}$, while for self-attention heads,
$\valpha_{t,v}^{(l,h)}=[\alpha_{t,1}^{(l,h)},\dots,\alpha_{t,|\vx_v|}^{(l,h)}]$, where $|\vx_v|$ denotes the number of visual tokens or patches. 
We then define the attention-based visual grounding score for token $y_t$ at layer $l$ as
\begin{equation}
    a_t^{(l)}(\vy,\vx)
    =
    \frac{1}{H}
    \sum_{h=1}^{H}
    \sum_{i=1}^{|\vx_v|}
    \bigl[\valpha_{t,v}^{(l,h)}\bigr]_i,
\end{equation}
where $H$ is the number of attention heads that interact with the visual content. 
A higher value of $a_t^{(l)}(\vy,\vx)$ indicates stronger dependency on the visual modality, whereas a lower value suggests that visual information contributes minimally to the next-token prediction at that layer.
As for the distribution-based score, we normalize these raw attention scores across all generated tokens to obtain a relative token-level importance score $S_A^{(l)}(y_t,\vx)$, where 
\begin{equation}
    S_{A}(\vy_t,\vx)=\frac{a_t(\vy_t,\vx)}{\sum_{i=1}^Ta_t(\vy_i,\vx)}.
    \label{def:S_A}
\end{equation}
This normalized score captures the relative contribution of token $\vy_t$ to the overall visual attention mass in the generated answer.

\subsection{\method}
\label{sec:method}
Our approach leverages visual grounding scores defined in Section~\ref{sec:vgs} and combines them with token-level uncertainty scores. 
Our approach assumes that tokens more strongly grounded in visual content contribute more significantly to uncertainty estimation, as they are more likely to support factual claims in the generated output (see Section~\ref{sec:experiments}).

\paragraph{Visual-grounding token UQ.}

We define \method~as a UQ framework for LVLMs that weights token-level uncertainty scores with visual grounding scores.
Specifically, given a multimodal input $\vx$ and a generated answer $\vy=\bigl[\vy_t\bigr]_{t=1}^T$, we define a visual-grounded uncertainty score by combining the distribution-based visual grounding score $S_{\mathrm{JSD}}(\cdot,\cdot)$ (\eqref{def:js_raw_score}) and the attention-based visual grounding score $S_A^{(l)}(\cdot,\cdot)$ (\eqref{def:S_A}) with the token entropy $\mathcal{H}(\vy_t \mid \vx)$ (\eqref{def:token_entropy}), and summing over all generated tokens:
\begin{multline}
    \label{eq_vgtuq}
    \operatorname{VIG\text{-}TUQ}(\vx)
    =\\
    \sum_{t=1}^{T}
    \Bigl[
        \alpha_{\mathrm{JSD}} S_{\mathrm{JSD}}(\vy_t, \vx)
        +
        \alpha_A S_A^{(l)}(\vy_t,\vx)
    \Bigr] 
    \mathcal{H}(\vy_t \mid \vx),
\end{multline}

where $\alpha_{\mathrm{JSD}}>0$ and $\alpha_A>0$ are weighting hyperparameters, and $l$ denotes the layer used to compute $S_A^{(l)}(\cdot,\cdot)$.
\joseph{Importantly, the grounding scores do not represent uncertainty themselves: they determine how strongly each token-level entropy contributes to the uncertainty of the full generated sequence. Tokens that are weakly grounded in the visual input contribute less to the final uncertainty score, while strongly grounded tokens contribute more through their entropy. The sequence-level uncertainty is therefore obtained by combining the entropy of all generated tokens according to their weighted by their degree of visual grounding.}%

\section{Experiments}
\label{sec:experiments}

\subsection{Experimental setup}

\paragraph{Datasets and models.}
We evaluate our method across a range of visual question answering (VQA) tasks, datasets, and model architectures. 
Specifically, we conduct experiments on four benchmark datasets: \textbf{ADVQA}~\citep{li2021adversarial}, \textbf{VQARAD}~\citep{lau2018dataset}, \textbf{OKVQA}~\citep{marino2019ok}, and \textbf{VizWiz}~\citep{gurari2018vizwiz}. 
Our evaluation includes several recent LVLM architectures, namely \texttt{Qwen2.5-VL-3B} and \texttt{Qwen2.5-VL-7B}~\citep{bai2025qwen2}, \texttt{Qwen3.5-2B}~\citep{qwen3.5}, \texttt{llava1.5-7B}, \texttt{llava1.5-13B}~\citep{liu2024improved}, \texttt{idefics-9B}~\citep{laurenccon2023obelics}  and \texttt{idefics2-8B}~\citep{laurenccon2023obelics}. Additional details regarding the datasets, models, and experimental setup are provided in Appendix~\ref{sec:Experimental_Protocol}.

\paragraph{Evaluation metrics.}
Following prior work~\citep{kuhnsemantic, farquhar2024detecting, janiak2025, lau2025uncertainty}, we evaluate uncertainty estimation methods using the AUROC metric~\citep{hendrycks2016baseline} and ECE~\citep{guo2017calibration}. 
AUROC (Area Under the Receiver Operating Curve) measures the ability of an uncertainty score to distinguish between correct and incorrect model predictions. 
Higher values indicate better discrimination, with a perfect uncertainty estimator achieving a score of $1$, while a random uncertainty measure yields an expected score of $0.5$. ECE (Expected Calibration Error) quantifies how well predicted probabilities align with empirical accuracies, where lower values indicate better calibration.
Additional details on the evaluation metrics are provided in Appendix~\ref{sec:Experimental_Protocol}.

\paragraph{Ground truth label.}
To evaluate the correctness of generated answers, we adopt an \emph{LLM-as-a-Judge} approach, following the recommendation of~\citep{janiak2025}. 
For each generated answer, the dataset provides a set of valid ground-truth responses. 
We use \texttt{Llama-3.1-8B} to determine whether the generated answer is semantically consistent with any of the provided ground-truth answers. 
This setup has been shown to be reliable and to exhibit strong agreement with human judgments~\citep{janiak2025illusion}.

\paragraph{Baselines.}
We compare our method against several widely used token-level (TL) uncertainty baselines as well as two sample-diversity approaches, that are semantic-level (SL). 
For TL methods, we include Token Entropy (\eqref{def:token_entropy}), Log-Perplexity (\eqref{def:LP}), Max-Prob~\citep{Fomicheva2020, malinin2021}, CCP~\ref{def:CCP}, and Token-SAR (\eqref{def:SAR}). 
We also include results of SL methods like the Semantic Entropy (SE)~\citep{kuhnsemantic} and the Kernel Language Entropy (KLE) with a Matern kernel~\citep{nikitin2024kernel}.
\joseph{Importantly, our primary comparison is with token-level methods, which estimate uncertainty from a single generated answer. 
We also report SE and KLE as stronger but more expensive reference methods, since they require multiple generated answers to estimate the uncertainty. Hyperparameter details are provided in Appendix~\ref{app_hyperparam}.}

\subsection{Results}
\paragraph{\textbf{Overall performance comparison.}} 
Tables~\ref{tab:per_data_set_results} and~\ref{tab:res_per_model} report the performance of the evaluated uncertainty estimation methods. 
Table~\ref{tab:per_data_set_results} presents AUROC scores aggregated across LVLM architectures for each dataset, while Table~\ref{tab:res_per_model} reports results aggregated across datasets for each model architecture.
\joseph{Across all datasets, \method~achieves the best or tied-best AUROC among token-level methods, improving over Token-Entropy by up to 4.2 AUROC points on VQARAD.
Remarkably, despite using only a single generated answer, \method~also remains competitive with the more expensive semantic-level methods: it outperforms SE on VizWiz and VQARAD, nearly matches it on ADVQA, and even outperforms KLE on VQARAD.}
Consequently, they incur a significantly higher inference cost compared to token-level approaches. 
\joseph{Finally, we compute standard deviations of our AUROCs in Tab \ref{tab:per_data_set_results}, \ref{tab:global_hparams} and Tab \ref{tab:fuull_results} that confirm the statistical significance of our results.}
\joseph{Overall, these results show that \method~improves token-level UQ consistently while retaining a computational cost close to lightweight single-answer baselines, and approaches the performance of more expensive semantic-level methods.}

\begin{table*}[t]
\caption{\textbf{\method ~%
improves AUROC compared to all token-level (TL) baselines and remains competitive with semantic-level (SL) baselines.} Evaluation metrics for datasets averaged across all LVLM architectures. Values in parentheses are 95\% confidence intervals.
\textbf{Bold} values indicates best performance, \underline{underline} values indicate the second-best among TL.}
\label{tab:per_data_set_results}
\centering
\resizebox{0.9\textwidth}{!}{
\begin{tabular}{cc|cccccccc@{\hskip 1.2em}|@{\hskip 1.2em}c}
\toprule
&\multicolumn{1}{l}{} & \multicolumn{2}{c}{\textbf{OKVQA}} & \multicolumn{2}{c}{\textbf{VIZWIZ}} & \multicolumn{2}{c}{\textbf{VQARAD}} & \multicolumn{2}{c}{\textbf{ADVQA}} & \textbf{Compute} \\

&\multicolumn{1}{l}{} & AUROC$\uparrow$ & ECE $\downarrow$ & AUROC$\uparrow$ & ECE$\downarrow$ & AUROC$\uparrow$ & ECE$\downarrow$ & AUROC$\uparrow$ & ECE$\downarrow$ &  s \\
\midrule
\multirow{7}{*}{\rotatebox{90}{TL}}
& \textit{Token-Sar} & 0.591 (0.008) & 0.137 (0.007) & 0.588 (0.012) & 0.126 (0.010) & 0.616 (0.027) & 0.183 (0.021) & 0.564 (0.015) & 0.173 (0.013) & 1.56 \\
 & \textit{CCP} & 0.625 (0.008) & 0.112 (0.007) & 0.640 (0.012) & 0.101 (0.010) & 0.621 (0.027) & \textbf{0.163} (0.021) & \textbf{0.631} (0.015) & \textbf{0.126} (0.013) & 3.35 \\
 & \textit{Max-Prob} & 0.621 (0.008) & 0.125 (0.007) & 0.641 (0.012) & 0.119 (0.010) & 0.626 (0.027) & 0.194 (0.021) & 0.621 (0.015) & 0.180 (0.013) & 1.02\\
 & \textit{Token-Entropy} & 0.627 (0.008) & 0.109 (0.007) & 0.650 (0.012) & 0.093 (0.010) & 0.642 (0.027) & 0.200 (0.021) & 0.612 (0.015) & 0.168 (0.013) & 1.01\\
 & \textit{\method}-A (ours) & \underline{0.648} (0.008) & \underline{0.100} (0.007) & \underline{0.676} (0.012) & \textbf{0.086} (0.010) & \underline{0.670} (0.027) & 0.194 (0.021) & \underline{0.629} (0.015) & 0.161 (0.013) & 1.21\\
 & \textit{\method}-JSD (ours) & 0.628 (0.008) & \textbf{0.078} (0.007) & 0.664 (0.012) & 0.097 (0.010) & 0.638 (0.027) & 0.210 (0.021) & 0.605 (0.015) & 0.164 (0.013) & 1.32 \\
 \rowcolor{green!20} & \textit{\method} (ours) & \textbf{0.653} (0.008) & \underline{0.100} (0.007) & \textbf{0.679} (0.012) & \underline{0.088} (0.010) & \textbf{0.684} (0.027) & \underline{0.180} (0.021) & \textbf{0.631} (0.015) & \underline{0.159} (0.013) & 1.37 \\
 \midrule
\multirow{2}{*}{\rotatebox{90}{SL}}
 
& \textit{SE} \joseph{(Gold)} & 0.664 (0.008) & 0.074 (0.007) & 0.649 (0.012) & 0.061 (0.010) & 0.652 (0.027) & 0.174 (0.021) & 0.632 (0.015) & 0.194 (0.013) & 4.69 \\
 & \textit{KLE} \joseph{(Gold)} & 0.690 (0.008) & 0.072 (0.007) & 0.682 (0.012) & 0.063 (0.010) & 0.656 (0.027) & 0.174 (0.021) & 0.646 (0.015) & 0.192 (0.013) & 5.68 \\
\midrule
\end{tabular}

}
\caption*{Compute is the mean inference time in second for generation + uncertainty quantification.}
\end{table*}

\begin{table*}[t]
\caption{\textbf{\method ~%
improves AUROC compared to all token-level (TL) baselines and remains competitive with semantic-level (SL) baselines.} Evaluation metrics for LVLM architectures averaged across all vision datasets.
\textbf{Bold} values indicates best performance. \underline{underline} values indicate the second-best among TL.}
\label{tab:res_per_model}
\resizebox{0.9\textwidth}{!}{
\begin{tabular}{cc|ccccccccccccccc}
\toprule
&& \multicolumn{2}{c}{\texttt{Qwen2.5-VL-3B}} & \multicolumn{2}{c}{\texttt{Qwen2.5-VL-7B}} & \multicolumn{2}{c}{\texttt{llava-1.5-13B}} & \multicolumn{2}{c}{\texttt{llava-1.5-7B}} & \multicolumn{2}{c}{\texttt{idefics2-8B}} & \multicolumn{2}{c}{\texttt{idefics-9B}} & \multicolumn{2}{c}{\texttt{Qwen3.5-2B}} \\

& Method &  AUROC$\uparrow$ & ECE$\downarrow$ & AUROC$\uparrow$ & ECE$\downarrow$ & AUROC$\uparrow$ & ECE$\downarrow$ & AUROC$\uparrow$ & ECE$\downarrow$ & AUROC$\uparrow$ & ECE$\downarrow$ & AUROC$\uparrow$ & ECE$\downarrow$ & AUROC$\uparrow$ & ECE$\downarrow$ \\
\midrule

\multirow{6}{*}{\rotatebox{90}{TL}}
& \textit{Token-Sar} & 0.554 & 0.175 & 0.588 & 0.183 & 0.639 & 0.136  & 0.609 & 0.136 & 0.599 & 0.153 & 0.550 & 0.170 & 0.587 & 0.130 \\
 & \textit{CCP} & 0.644 & 0.120 & 0.620 & \underline{0.148} & 0.627 & 0.137 & 0.624 & \textbf{0.120} & 0.627 & \textbf{0.137} & \underline{0.659} & \textbf{0.118} & 0.604 & \textbf{0.099} \\
 & \textit{Max-Prob} & 0.643 & 0.150 & 0.616 & 0.172 & 0.625 & 0.163 & 0.623 & 0.159 & 0.617 & 0.162 & 0.656 & 0.164 & 0.611 & 0.112 \\
 & \textit{Token-Entropy} & 0.651 & 0.126 & 0.622 & 0.164 & 0.650 & 0.140 & 0.632 & 0.133 & 0.628 & 0.172 & 0.629 & 0.154 & 0.616 & 0.109 \\
 & \textit{\method}-A (ours) & \underline{0.689} &\underline{ 0.116} & \underline{0.651} & 0.154 & \underline{0.660} & 0.147 & 0.640 & 0.139 & \underline{0.659} & \underline{0.139} & \textbf{0.660} & \underline{0.144} & \underline{0.632} & 0.107 \\
 & \textit{\method}-JSD (ours) & 0.652 & \textbf{0.112} & 0.611 & \textbf{0.142} & 0.657 & \textbf{0.123} & \underline{0.652} & \underline{0.124} & 0.651 & 0.150 & 0.600 & 0.197 & 0.614 & 0.113 \\
 
 \rowcolor{green!20} & \textit{\method} (ours) & \textbf{0.691} & \textbf{0.112} & \textbf{0.653} & 0.153 & \textbf{0.671} & \underline{0.135} & \textbf{0.663} & 0.126 & \textbf{0.661} & 0.142 & \textbf{0.660} & \underline{0.144} & \textbf{0.633} & \underline{0.106} \\

 \midrule

\multirow{2}{*}{\rotatebox{90}{SL}}
& \textit{SE} \joseph{(Gold)} & 0.662 & 0.114 & 0.651 & 0.148 & 0.668 & 0.127 & 0.680 & 0.121 & 0.655 & 0.125 & 0.674 & 0.121 & 0.555 & 0.123 \\
 & \textit{KLE} \joseph{(Gold)} & 0.673 & 0.113 & 0.658 & 0.147 & 0.683 & 0.127 & 0.686 & 0.121 & 0.668 & 0.124 & 0.687 & 0.120 & 0.626 & 0.127 \\

\midrule

\end{tabular}

}
\caption*{\footnotesize \textit{\method~} is as described in Eq\ref{eq_vgtuq}, \textit{{\method}-A} and \textit{{\method}-JSD} are variants using only one of the visual grounding criteria. }
\end{table*}

\begin{figure*}[t!]
    \centering
    \begin{center}
        \includegraphics[width=0.7\linewidth]{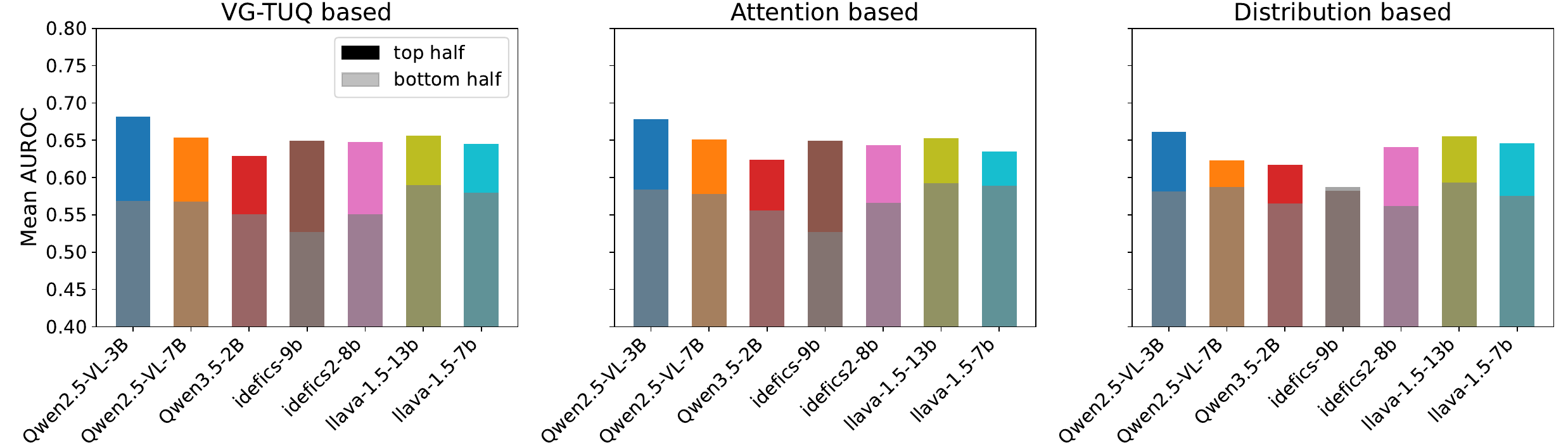}
    \end{center}
    \caption{\textbf{Visual grounding scores help identify the tokens most relevant for uncertainty estimation.}
AUROC performance reported for the mean token entropy of the top half vs. bottom half of generated tokens, ranked by a given selector, across different LVLM architectures.
Tokens are ranked according to the distribution-based score (\eqref{def:js_raw_score}), the attention-based score (\eqref{def:S_A}), or their weighted combination as used in \method~ (details in Appendix~\ref{sec:token_selection}).
Colored bars correspond to the top half of tokens and gray bars to the bottom half.
All results are averaged across all datasets.}
    \label{fig:token_selection}
\end{figure*}

\paragraph{Visual grounding \& informative tokens.}
\label{sub_sec:visual_grounding}

\joseph{
Our approach is based on the assumption that visual grounding scores help identifying the most relevant tokens for uncertainty estimation.
We further hypothesize that the effectiveness of these scores is model-dependent and that they capture distinct aspects of visual reasoning.
To test this hypothesis, we split generated tokens into a top half and a bottom half according to a given selector, and compute the mean token entropy within each half.
Tokens are ranked using either the distribution-based score (\eqref{def:js_raw_score}), the attention-based score (\eqref{def:S_A}), or their weighted combination (\eqref{eq_vgtuq}, as used in \method). Importantly, visual grounding is used only to rank tokens in this experiment and does not enter the uncertainty score itself. While Token-SAR estimates token importance from semantic relevance using an external semantic model, and CCP relies on an external NLI model to assess semantic consistency among token alternatives, our results show that selecting tokens according to their visual grounding yields more informative token entropies for sequence-level uncertainty estimation. This indicates that, in LVLMs, visual grounding provides a more task-relevant and trustworthy notion of token importance for uncertainty estimation, while avoiding the additional external models required by Token-SAR and CCP.
Results presented in Figure~\ref{fig:token_selection} lead to three main observations.
(1) For all three selectors, the top half consistently yields higher mean AUROC than the bottom half across models and datasets, confirming that visual grounding scores identify tokens carrying more informative uncertainty signal;
(2) The attention-based and distribution-based criteria perform better on different models and datasets, which confirms their complementary nature and motivates their combination in \method;
(3) The weighted combination used in \method~ matches or exceeds the top-half performance of either individual criterion, showing that combining both signals yields a more robust token selector across models.
Overall, these findings suggest that uncertainty is not uniformly distributed across the generated sequence and that only a subset of tokens carries the most informative signals for uncertainty quantification.
Additional experiments about the validity of our grounding score are in Appendix~\ref{sec:token_selection}.
}

\paragraph{\textbf{Computational cost.}} 
We evaluate the efficiency of \method~ by comparing its average inference time with existing token-level UQ methods on OKVQA. 
Table~ \ref{tab:per_data_set_results} and \ref{tab:speed_test} reports the average runtime for answer generation and uncertainty estimation under identical backbone models and hardware settings. 
Since all token-level methods share the same decoding procedure, generation time remains constant and any runtime differences arise only from the post-hoc UQ computation.
Unlike CCP and Token-SAR, which depend on external models and thus introduce additional computational overhead, \method~ is more efficient because it leverages only token probabilities and attention weights that are computed by the generation model itself. 
Although lightweight baselines such as Token-Entropy, Log-Perplexity, and Max-Prob remain slightly faster, \method~ incurs only minimal overhead while being substantially more efficient than other token-level UQ approaches.

\joseph{\subsection{Global Model-Level Hyperparameters}}

\joseph{Most UQ methods require some hyperparameter tuning. In our comparison, only simple baselines such as Token-Entropy and Max-Prob are parameter-free, while methods such as CCP, Token-SAR, SE, and KLE also require selecting thresholds, relevance parameters, or other settings. VIG-TUQ therefore follows the same general setting. In the main experiments, we tune its hyperparameters for each model--dataset pair using training data. However, for easier deployment, it is useful to know whether one configuration can be shared across datasets. We therefore evaluate a fixed configuration per model, shared across all four datasets.}

\joseph{Table~\ref{tab:global_hparams} reports AUROC values as bootstrapped means (standard deviations) over 1{,}000 resamples, macro-averaged across the four datasets (OKVQA, VIZWIZ, VQARAD, ADVQA). We compare three settings: (i) token entropy, (ii) VIG-TUQ (model only), using a single hyperparameter configuration shared across datasets (model-level), and (iii) VIG-TUQ (model/data), using per-dataset tuned hyperparameters.}

\begin{table}[t]
    \centering
    
    \caption{AUROC (bootstrapped mean, std. dev. over 1{,}000 resamples) macro-averaged across OKVQA, VIZWIZ, VQARAD, and ADVQA, comparing Token-Entropy, VIG-TUQ with shared model-level hyperparameters, and VIG-TUQ with per-dataset tuned hyperparameters.}
    
    \label{tab:global_hparams}
    \resizebox{0.45\textwidth}{!}{
    
    \begin{tabular}{lccc}
        \toprule
        \textbf{Model} & \textbf{Token-Entropy} & \textbf{VIG-TUQ (model only)} & \textbf{VIG-TUQ (model/data)} \\
        & AUROC$\uparrow$  & AUROC$\uparrow$ & AUROC$\uparrow$ \\
        \midrule
        \texttt{Qwen2.5-VL-3B}  & 0.651 (0.015) &\underline{ 0.690} (0.015) & \textbf{0.691} (0.015) \\
        \texttt{Qwen2.5-VL-7B}  & 0.622 (0.017) & \underline{0.642} (0.017) & \textbf{0.653} (0.016) \\
        \texttt{llava-1.5-13b}       & 0.650 (0.016) & \underline{0.668} (0.015) & \textbf{0.671} (0.011) \\
        \texttt{llava-1.5-7b}        & 0.632 (0.016) & \underline{0.655} (0.016) & \textbf{0.663} (0.016) \\
        \texttt{idefics2-8b}     & 0.627 (0.016) & \underline{0.647} (0.016) & \textbf{0.661} (0.015) \\
        \texttt{idefics-9b}    & 0.629 (0.016) & \underline{0.658} (0.015) & \textbf{0.660} (0.015) \\
        \texttt{Qwen3.5-2B}               & 0.616 (0.015) & \underline{0.627} (0.015) & \textbf{0.633} (0.015) \\
        \bottomrule
    \end{tabular}
    }
\end{table}
\joseph{For every model, the gap between VIG-TUQ (model only) and Token-Entropy exceeds the standard deviation of either estimate, indicating that the observed gains are statistically meaningful rather than noise. Notably, even when hyperparameters are shared across all four datasets rather than tuned per dataset, VIG-TUQ still yields a significant and consistent improvement over the baseline for every model. These results indicate that a single, easy-to-deploy hyperparameter setting per model is sufficient to achieve strong and statistically robust gains, supporting the practicality of our approach beyond the per-dataset tuning setup used in the main experiments.}

\section{Conclusion}

In this paper, we present a token-level analysis of how reliance on visual evidence in LVMs affects uncertainty quantification.
Our findings indicate that visual information helps identify the most relevant token for uncertainty quantification and underscore the importance of combining multiple visual grounding signals, as their effectiveness can vary across models and capture complementary aspects of visual reasoning.
Building on these insights, we introduce Visual-Grounded Token UQ (VIG-TUQ), a training-free and computational efficient framework that explicitly incorporates visual grounding into uncertainty estimation by weighting token-level language uncertainty with visual grounding scores.
Experiments on multiple datasets and across diverse LVLM architectures (including early-fusion, late-fusion, and native-fusion models) indicate that our method often improves upon existing token-level uncertainty approaches.
We hope our work will encourage future research on vision-aware and fine-grained uncertainty quantification for LVLMs.

\textbf{Limitations.} While our approach improves upon existing token-level uncertainty methods, semantic uncertainty methods can still perform better in some cases, but they are more costly. Combining them with visual grounding could be explored in future work.

\clearpage

{
    \small
    \bibliographystyle{ieeenat_fullname}
    \bibliography{main}

@String(CVPR= {IEEE Conf. Comput. Vis. Pattern Recog.})

@String(ICCV= {Int. Conf. Comput. Vis.})

@String(ECCV= {Eur. Conf. Comput. Vis.})

@String(ICLR = {Int. Conf. Learn. Represent.})

@String(CVPR  = {CVPR})

@String(ICCV  = {ICCV})

@String(ECCV  = {ECCV})

@String(ICLR  = {ICLR})

@inproceedings{
long2026understanding,
title={Understanding Language Prior of {LVLM}s by Contrasting Chain-of-Embedding},
author={Lin Long and Changdae Oh and Seongheon Park and Sharon Li},
booktitle={The Fourteenth International Conference on Learning Representations},
year={2026},
url={https://openreview.net/forum?id=wnTnwWKu72}
}

@inproceedings{kim2025,
    title = "Detecting {LLM} Hallucination Through Layer-wise Information Deficiency: Analysis of Ambiguous Prompts and Unanswerable Questions",
    author = "Kim, Hazel  and
      Lamb, Tom A.  and
      Bibi, Adel  and
      Torr, Philip  and
      Gal, Yarin",
    editor = "Christodoulopoulos, Christos  and
      Chakraborty, Tanmoy  and
      Rose, Carolyn  and
      Peng, Violet",
    booktitle = "Proceedings of the 2025 Conference on Empirical Methods in Natural Language Processing",
    month = nov,
    year = "2025",
    address = "Suzhou, China",
    publisher = "Association for Computational Linguistics",
    url = "https://aclanthology.org/2025.emnlp-main.1644/",
    doi = "10.18653/v1/2025.emnlp-main.1644",
    pages = "32310--32322",
    ISBN = "979-8-89176-332-6",
}

@inproceedings{leeetal2025,
    title = "{VL}ind-Bench: Measuring Language Priors in Large Vision-Language Models",
    author = "Lee, Kang-il  and
      Kim, Minbeom  and
      Yoon, Seunghyun  and
      Kim, Minsung  and
      Lee, Dongryeol  and
      Koh, Hyukhun  and
      Jung, Kyomin",
    editor = "Chiruzzo, Luis  and
      Ritter, Alan  and
      Wang, Lu",
    booktitle = "Findings of the Association for Computational Linguistics: NAACL 2025",
    month = apr,
    year = "2025",
    address = "Albuquerque, New Mexico",
    publisher = "Association for Computational Linguistics",
    url = "https://aclanthology.org/2025.findings-naacl.231/",
    doi = "10.18653/v1/2025.findings-naacl.231",
    pages = "4129--4144",
    ISBN = "979-8-89176-195-7",
}

@inproceedings{
luo2025probing,
title={Probing Visual Language Priors in {VLM}s},
author={Tiange Luo and Ang Cao and Gunhee Lee and Justin Johnson and Honglak Lee},
booktitle={Forty-second International Conference on Machine Learning},
year={2025},
url={https://openreview.net/forum?id=bhTBirS0qi}
}

@inproceedings{fu2025,
author="Fu, Xingyu
and Hu, Yushi
and Li, Bangzheng
and Feng, Yu
and Wang, Haoyu
and Lin, Xudong
and Roth, Dan
and Smith, Noah A.
and Ma, Wei-Chiu
and Krishna, Ranjay",
editor="Leonardis, Ale{\v{s}}
and Ricci, Elisa
and Roth, Stefan
and Russakovsky, Olga
and Sattler, Torsten
and Varol, G{\"u}l",
title="BLINK: Multimodal Large Language Models Can See but Not Perceive",
booktitle="Computer Vision -- ECCV 2024",
year="2025",
publisher="Springer Nature Switzerland",
address="Cham",
pages="148--166",
}

@article{kadavath2022language,
  title={Language models (mostly) know what they know},
  author={Kadavath, Saurav and Conerly, Tom and Askell, Amanda and Henighan, Tom and Drain, Dawn and Perez, Ethan and Schiefer, Nicholas and Hatfield-Dodds, Zac and DasSarma, Nova and Tran-Johnson, Eli and others},
  journal={arXiv preprint arXiv:2207.05221},
  year={2022}
}

@inproceedings{ji2025calibrating,
    title = "Calibrating Verbal Uncertainty as a Linear Feature to Reduce Hallucinations",
    author = "Ji, Ziwei  and
      Yu, Lei  and
      Koishekenov, Yeskendir  and
      Bang, Yejin  and
      Hartshorn, Anthony  and
      Schelten, Alan  and
      Zhang, Cheng  and
      Fung, Pascale  and
      Cancedda, Nicola",
    editor = "Christodoulopoulos, Christos  and
      Chakraborty, Tanmoy  and
      Rose, Carolyn  and
      Peng, Violet",
    booktitle = "Proceedings of the 2025 Conference on Empirical Methods in Natural Language Processing",
    month = nov,
    year = "2025",
    address = "Suzhou, China",
    publisher = "Association for Computational Linguistics",
    url = "https://aclanthology.org/2025.emnlp-main.187/",
    doi = "10.18653/v1/2025.emnlp-main.187",
    pages = "3769--3793",
    ISBN = "979-8-89176-332-6",
}

@inproceedings{
xiong2024can,
title={Can {LLM}s Express Their Uncertainty? An Empirical Evaluation of Confidence Elicitation in {LLM}s},
author={Miao Xiong and Zhiyuan Hu and Xinyang Lu and YIFEI LI and Jie Fu and Junxian He and Bryan Hooi},
booktitle={The Twelfth International Conference on Learning Representations},
year={2024},
url={https://openreview.net/forum?id=gjeQKFxFpZ}
}

@article{yona2024can,
  title={Can large language models faithfully express their intrinsic uncertainty in words?},
  author={Yona, Gal and Aharoni, Roee and Geva, Mor},
  journal={arXiv preprint arXiv:2405.16908},
  year={2024}
}

@article{han2025multimodal,
  title={Multimodal fusion and vision-language models: A survey for robot vision},
  author={Han, Xiaofeng and Chen, Shunpeng and Fu, Zenghuang and Feng, Zhe and Fan, Lue and An, Dong and Wang, Changwei and Guo, Li and Meng, Weiliang and Zhang, Xiaopeng and others},
  journal={Information Fusion},
  pages={103652},
  year={2025},
  publisher={Elsevier}
}

@inproceedings{
malinin2021uncertainty,
title={Uncertainty Estimation in Autoregressive Structured Prediction},
author={Andrey Malinin and Mark Gales},
booktitle={International Conference on Learning Representations},
year={2021},
url={https://openreview.net/forum?id=jN5y-zb5Q7m}
}

@misc{kang2025uncertainty,
      title={Uncertainty Quantification for Hallucination Detection in Large Language Models: Foundations, Methodology, and Future Directions}, 
      author={Sungmin Kang and Yavuz Faruk Bakman and Duygu Nur Yaldiz and Baturalp Buyukates and Salman Avestimehr},
      year={2025},
      eprint={2510.12040},
      archivePrefix={arXiv},
      primaryClass={cs.CL},
      url={https://arxiv.org/abs/2510.12040}, 
}

@misc{zhang2024vluncertainty,
      title={VL-Uncertainty: Detecting Hallucination in Large Vision-Language Model via Uncertainty Estimation}, 
      author={Ruiyang Zhang and Hu Zhang and Zhedong Zheng},
      year={2024},
      eprint={2411.11919},
      archivePrefix={arXiv},
      primaryClass={cs.CV},
      url={https://arxiv.org/abs/2411.11919}, 
}

@article{jiang2024evaluating,
  title={Evaluating general vision-language models for clinical medicine},
  author={Jiang, Yixing and Omiye, Jesutofunmi A and Zakka, Cyril and Moor, Michael and Gui, Haiwen and Alipour, Shayan and Mousavi, Seyed Shahabeddin and Chen, Jonathan H and Rajpurkar, Pranav and Daneshjou, Roxana},
  journal={MedRxiv},
  pages={2024--04},
  year={2024},
  publisher={Cold Spring Harbor Laboratory Press}
}

@inproceedings{gao2025survey,
  title={A survey for foundation models in autonomous driving},
  author={Gao, Haoxiang and Wang, Zhongruo and Li, Yaqian and Long, Kaiwen and Yang, Ming and Shen, Yiqing},
  booktitle={2025 6th International Conference on Computer Vision and Data Mining (ICCVDM)},
  pages={63--71},
  year={2025},
  organization={IEEE}
}

@misc{openai2025gptoss,
      title={gpt-oss-120b and gpt-oss-20b Model Card}, 
      author={OpenAI},
      year={2025},
      eprint={2508.10925},
      archivePrefix={arXiv},
      primaryClass={cs.CL},
      url={https://arxiv.org/abs/2508.10925}, 
}

@article{Fomicheva2020,
    author = {Fomicheva, Marina and Sun, Shuo and Yankovskaya, Lisa and Blain, Frédéric and Guzmán, Francisco and Fishel, Mark and Aletras, Nikolaos and Chaudhary, Vishrav and Specia, Lucia},
    title = {Unsupervised Quality Estimation for Neural Machine Translation},
    journal = {Transactions of the Association for Computational Linguistics},
    volume = {8},
    pages = {539-555},
    year = {2020},
    month = {09},
    issn = {2307-387X},
    doi = {10.1162/tacl_a_00330},
    url = {https://doi.org/10.1162/tacl_a_00330},
    eprint = {https://direct.mit.edu/tacl/article-pdf/doi/10.1162/tacl_a_00330/1923296/tacl_a_00330.pdf},
}

@inproceedings{
malinin2021,
title={Uncertainty Estimation in Autoregressive Structured Prediction},
author={Andrey Malinin and Mark Gales},
booktitle={ICLR},
year={2021},
}

@article{alayrac2022flamingo,
  title={Flamingo: a visual language model for few-shot learning},
  author={Alayrac, Jean-Baptiste and Donahue, Jeff and Luc, Pauline and Miech, Antoine and Barr, Iain and Hasson, Yana and Lenc, Karel and Mensch, Arthur and Millican, Katherine and Reynolds, Malcolm and others},
  journal={Advances in neural information processing systems},
  volume={35},
  pages={23716--23736},
  year={2022}
}

@article{laurenccon2023obelics,
  title={Obelics: An open web-scale filtered dataset of interleaved image-text documents},
  author={Lauren{\c{c}}on, Hugo and Saulnier, Lucile and Tronchon, L{\'e}o and Bekman, Stas and Singh, Amanpreet and Lozhkov, Anton and Wang, Thomas and Karamcheti, Siddharth and Rush, Alexander and Kiela, Douwe and others},
  journal={Advances in Neural Information Processing Systems},
  volume={36},
  pages={71683--71702},
  year={2023}
}

@misc{openai2024gpt4,
      title={GPT-4 Technical Report}, 
      author={OpenAI},
      year={2024},
      eprint={2303.08774},
      archivePrefix={arXiv},
      primaryClass={cs.CL},
      url={https://arxiv.org/abs/2303.08774}, 
}

@misc{
bai2024qwenvl,
title={Qwen-{VL}: A Versatile Vision-Language Model for Understanding, Localization, Text Reading, and Beyond},
author={Jinze Bai and Shuai Bai and Shusheng Yang and Shijie Wang and Sinan Tan and Peng Wang and Junyang Lin and Chang Zhou and Jingren Zhou},
year={2024},
url={https://openreview.net/forum?id=qrGjFJVl3m}
}

@article{hoche2025improving,
  title={Improving Semantic Uncertainty Quantification in LVLMs with Semantic Gaussian Processes},
  author={Hoche, Joseph and Bursuc, Andrei and Brellmann, David and Louppe, Gilles and Izmailov, Pavel and Yao, Angela and Franchi, Gianni},
  journal={arXiv preprint arXiv:2512.14177},
  year={2025}
}

@misc{
lau2025uncertainty,
title={Uncertainty Quantification for Multimodal Large Language Models with Coherence-adjusted Semantic Volume},
author={Gregory Kang Ruey Lau and Hieu Dao and Nicole Kan Hui Lin and Bryan Kian Hsiang Low},
year={2025},
url={https://openreview.net/forum?id=c9TWeKZQR4}
}

@misc{geminiteam2024,
      title={Gemini 1.5: Unlocking multimodal understanding across millions of tokens of context}, 
      author={Gemini},
year={2024},
      eprint={2403.05530},
      archivePrefix={arXiv},
      primaryClass={cs.CL},
      url={https://arxiv.org/abs/2403.05530}, 
}

@article{novikov2025alphaevolve,
  title={AlphaEvolve: A coding agent for scientific and algorithmic discovery},
  author={Novikov, Alexander and V{\~u}, Ng{\^a}n and Eisenberger, Marvin and Dupont, Emilien and Huang, Po-Sen and Wagner, Adam Zsolt and Shirobokov, Sergey and Kozlovskii, Borislav and Ruiz, Francisco JR and Mehrabian, Abbas and others},
  journal={arXiv preprint arXiv:2506.13131},
  year={2025}
}

@inproceedings{liu2023visual,
 author = {Liu, Haotian and Li, Chunyuan and Wu, Qingyang and Lee, Yong Jae},
 booktitle = {Advances in Neural Information Processing Systems},
 editor = {A. Oh and T. Naumann and A. Globerson and K. Saenko and M. Hardt and S. Levine},
 pages = {34892--34916},
 publisher = {Curran Associates, Inc.},
 title = {Visual Instruction Tuning},
 url = {https://proceedings.neurips.cc/paper_files/paper/2023/file/6dcf277ea32ce3288914faf369fe6de0-Paper-Conference.pdf},
 volume = {36},
 year = {2023}
}

@inproceedings{liu2024improved,
  title={Improved baselines with visual instruction tuning},
  author={Liu, Haotian and Li, Chunyuan and Li, Yuheng and Lee, Yong Jae},
  booktitle={CVPR},
  year={2024}
}

@article{liu2024survey,
  title={A survey on hallucination in large vision-language models},
  author={Liu, Hanchao and Xue, Wenyuan and Chen, Yifei and Chen, Dapeng and Zhao, Xiutian and Wang, Ke and Hou, Liping and Li, Rongjun and Peng, Wei},
  journal={arXiv preprint arXiv:2402.00253},
  year={2024}
}

@article{huang2025,
author = {Huang, Lei and Yu, Weijiang and Ma, Weitao and Zhong, Weihong and Feng, Zhangyin and Wang, Haotian and Chen, Qianglong and Peng, Weihua and Feng, Xiaocheng and Qin, Bing and Liu, Ting},
title = {A Survey on Hallucination in Large Language Models: Principles, Taxonomy, Challenges, and Open Questions},
year = {2025},
journal = {TIS},
}

@article{ji2023,
author = {Ji, Ziwei and Lee, Nayeon and Frieske, Rita and Yu, Tiezheng and Su, Dan and Xu, Yan and Ishii, Etsuko and Bang, Ye Jin and Madotto, Andrea and Fung, Pascale},
title = {Survey of Hallucination in Natural Language Generation},
year = {2023},
journal = {CS},
}

@article{guo2025deepseek,
  title={Deepseek-r1 incentivizes reasoning in llms through reinforcement learning},
  author={Guo, Daya and Yang, Dejian and Zhang, Haowei and Song, Junxiao and Wang, Peiyi and Zhu, Qihao and Xu, Runxin and Zhang, Ruoyu and Ma, Shirong and Bi, Xiao and others},
  journal={Nature},
  year={2025},
}

@article{farquhar2024detecting,
  title={Detecting hallucinations in large language models using semantic entropy},
  author={Farquhar, Sebastian and Kossen, Jannik and Kuhn, Lorenz and Gal, Yarin},
  journal={Nature},
  year={2024},
}

@inproceedings{nikitin2024kernel,
  title={Kernel language entropy: Fine-grained uncertainty quantification for llms from semantic similarities},
  author={Nikitin, Alexander and Kossen, Jannik and Gal, Yarin and Marttinen, Pekka},
  booktitle={NeurIPS},
  year={2024}
}

@article{grewal2024improving,
  title={Improving uncertainty quantification in large language models via semantic embeddings},
  author={Grewal, Yashvir S and Bonilla, Edwin V and Bui, Thang D},
  journal={arXiv preprint arXiv:2410.22685},
  year={2024}
}

@article{abdaljalil2025,
  title={Sindex: Semantic inconsistency index for hallucination detection in llms},
  author={Abdaljalil, Samir and Kurban, Hasan and Sharma, Parichit and Serpedin, Erchin and Atat, Rachad},
  journal={arXiv preprint arXiv:2503.05980},
  year={2025}
}

@inproceedings{
azaria2023the,
title={The Internal State of an {LLM} Knows When It's Lying},
author={Amos Azaria and Tom Mitchell},
booktitle={EMNLP},
year={2023},
}

@inproceedings{
chen2024,
title={{INSIDE}: {LLM}s' Internal States Retain the Power of Hallucination Detection},
author={Chao Chen and Kai Liu and Ze Chen and Yi Gu and Yue Wu and Mingyuan Tao and Zhihang Fu and Jieping Ye},
booktitle={ICLR},
year={2024},
}

@inproceedings{
orgad2025llms,
title={{LLM}s Know More Than They Show: On the Intrinsic Representation of {LLM} Hallucinations},
author={Hadas Orgad and Michael Toker and Zorik Gekhman and Roi Reichart and Idan Szpektor and Hadas Kotek and Yonatan Belinkov},
booktitle={ICLR},
year={2025},
}

@inproceedings{janiak2025,
    title = {The Illusion of Progress: Re-evaluating Hallucination Detection in {LLM}s},
    author = {Janiak, Denis  and Binkowski, Jakub  and Sawczyn, Albert  and    Gabrys, Bogdan  and Shwartz-Ziv, Ravid  and Kajdanowicz, Tomasz Jan},
    booktitle = {EMNLP},
    year = {2025}
}

@inproceedings{joocho2025,
    title = "Cleanse: Uncertainty Estimation Approach Using Clustering-based Semantic Consistency in {LLM}s",
    author = "Joo, Minsuh  and
      Cho, Hyunsoo",
    booktitle = "ACL Workshops",
    year = "2025",
}

@article{
lin2024generating,
title={Generating with Confidence: Uncertainty Quantification for Black-box Large Language Models},
author={Zhen Lin and Shubhendu Trivedi and Jimeng Sun},
journal={TMLR},
year={2024},
}

@inproceedings{
kuhnsemantic,
title={Semantic Uncertainty: Linguistic Invariances for Uncertainty Estimation in Natural Language Generation},
author={Lorenz Kuhn and Yarin Gal and Sebastian Farquhar},
booktitle={ICLR},
year={2023},
}

@InProceedings{gal2016dropout,
  title = 	 {Dropout as a Bayesian Approximation: Representing Model Uncertainty in Deep Learning},
  author = 	 {Gal, Yarin and Ghahramani, Zoubin},
  booktitle = 	 {ICML},
  year = 	 {2016},
}

@InProceedings{hernandez2015,
  title = 	 {Probabilistic Backpropagation for Scalable Learning of Bayesian Neural Networks},
  author = 	 {Hernandez-Lobato, Jose Miguel and Adams, Ryan},
  booktitle = 	 {ICML},
  year = 	 {2015},
}

@article{abdar2021,
title = {A review of uncertainty quantification in deep learning: Techniques, applications and challenges},
journal = {IF},
year = {2021},
author = {Moloud Abdar and Farhad Pourpanah and Sadiq Hussain and Dana Rezazadegan and Li Liu and Mohammad Ghavamzadeh and Paul Fieguth and Xiaochun Cao and Abbas Khosravi and U. Rajendra Acharya and Vladimir Makarenkov and Saeid Nahavandi},
}

@article{li2024multimodal,
  title={Multimodal alignment and fusion: A survey},
  author={Li, Songtao and Tang, Hao},
  journal={arXiv preprint arXiv:2411.17040},
  year={2024}
}

@article{an2025towards,
  title={Towards llm-centric multimodal fusion: A survey on integration strategies and techniques},
  author={An, Jisu and Lee, Junseok and Lee, Jeoungeun and Son, Yongseok},
  journal={arXiv preprint arXiv:2506.04788},
  year={2025}
}

@article{lau2018dataset,
  title={A dataset of clinically generated visual questions and answers about radiology images},
  author={Lau, Jason J and Gayen, Soumya and Ben Abacha, Asma and Demner-Fushman, Dina},
  journal={Scientific data},
  year={2018},
  }

@inproceedings{marino2019ok,
  title={Ok-vqa: A visual question answering benchmark requiring external knowledge},
  author={Marino, Kenneth and Rastegari, Mohammad and Farhadi, Ali and Mottaghi, Roozbeh},
  booktitle={CVPR},
  year={2019}
}

@inproceedings{li2021adversarial,
  title={Adversarial vqa: A new benchmark for evaluating the robustness of vqa models},
  author={Li, Linjie and Lei, Jie and Gan, Zhe and Liu, Jingjing},
  booktitle={ICCV},
  year={2021}
}

@inproceedings{gurari2018vizwiz,
  title={Vizwiz grand challenge: Answering visual questions from blind people},
  author={Gurari, Danna and Li, Qing and Stangl, Abigale J and Guo, Anhong and Lin, Chi and Grauman, Kristen and Luo, Jiebo and Bigham, Jeffrey P},
  booktitle={CVPR},
  year={2018}
}

@article{bai2025qwen2,
  title={Qwen2. 5-vl technical report},
  author={Bai, Shuai and Chen, Keqin and Liu, Xuejing and Wang, Jialin and Ge, Wenbin and Song, Sibo and Dang, Kai and Wang, Peng and Wang, Shijie and Tang, Jun and others},
  journal={arXiv preprint arXiv:2502.13923},
  year={2025}
}

@inproceedings{hendrycks2016baseline,
  title={A baseline for detecting misclassified and out-of-distribution examples in neural networks},
  author={Hendrycks, Dan and Gimpel, Kevin},
  booktitle={ICLR},
  year={2017}
}

@inproceedings{guo2017calibration,
  title={On calibration of modern neural networks},
  author={Guo, Chuan and Pleiss, Geoff and Sun, Yu and Weinberger, Kilian Q},
  booktitle={ICML},
  year={2017},
}

@inproceedings{kuhn2023semantic,
  title={Semantic uncertainty: Linguistic invariances for uncertainty estimation in natural language generation},
  author={Kuhn, Lorenz and Gal, Yarin and Farquhar, Sebastian},
  booktitle={ICLR},
  year={2023}
}

@article{fieback2024metatoken,
  title={Metatoken: Detecting hallucination in image descriptions by meta classification},
  author={Fieback, Laura and Spiegelberg, Jakob and Gottschalk, Hanno},
  journal={arXiv preprint arXiv:2405.19186},
  year={2024}
}

@inproceedings{janiak2025illusion,
  title={The illusion of progress: Re-evaluating hallucination detection in llms},
  author={Janiak, Denis and Binkowski, Jakub and Sawczyn, Albert and Gabrys, Bogdan and Shwartz-Ziv, Ravid and Kajdanowicz, Tomasz Jan},
  booktitle={Proceedings of the 2025 Conference on Empirical Methods in Natural Language Processing},
  pages={34716--34733},
  year={2025}
}

@inproceedings{fadeeva2024fact,
  title={Fact-checking the output of large language models via token-level uncertainty quantification},
  author={Fadeeva, Ekaterina and Rubashevskii, Aleksandr and Shelmanov, Artem and Petrakov, Sergey and Li, Haonan and Mubarak, Hamdy and Tsymbalov, Evgenii and Kuzmin, Gleb and Panchenko, Alexander and Baldwin, Timothy and others},
  booktitle={Findings of the Association for Computational Linguistics: ACL 2024},
  pages={9367--9385},
  year={2024}
}

@inproceedings{duan2024shifting,
  title={Shifting attention to relevance: Towards the predictive uncertainty quantification of free-form large language models},
  author={Duan, Jinhao and Cheng, Hao and Wang, Shiqi and Zavalny, Alex and Wang, Chenan and Xu, Renjing and Kailkhura, Bhavya and Xu, Kaidi},
  booktitle={Proceedings of the 62nd Annual Meeting of the Association for Computational Linguistics (Volume 1: Long Papers)},
  pages={5050--5063},
  year={2024}
}

@misc{qwen3.5,
    title  = {{Qwen3.5}: Towards Native Multimodal Agents},
    author = {{Qwen Team}},
    month  = {February},
    year   = {2026},
    url    = {https://qwen.ai/blog?id=qwen3.5}
}

@article{oquab2023dinov2,
  title={Dinov2: Learning robust visual features without supervision},
  author={Oquab, Maxime and Darcet, Timoth{\'e}e and Moutakanni, Th{\'e}o and Vo, Huy and Szafraniec, Marc and Khalidov, Vasil and Fernandez, Pierre and Haziza, Daniel and Massa, Francisco and El-Nouby, Alaaeldin and others},
  journal={arXiv preprint arXiv:2304.07193},
  year={2023}
}
}

\clearpage
\newpage
\appendix
\renewcommand{\thetable}{A.\arabic{table}}
\renewcommand{\thefigure}{A.\arabic{figure}}

\setcounter{page}{1}

\twocolumn[
\begin{center}
    \vspace*{0.3cm}
    {\Large\textbf{
    Leveraging Visual Signals for Robust Token-Level Uncertainty in Vision--Language Generation
    }}
    
    \vspace{0.2cm}
    
    {\large\textbf{-- Supplementary Material --}}
    
    \vspace{0.7cm}
\end{center}
]

\section*{Overview of the Supplementary Material}

The supplementary material provides additional analyses and complete experimental details supporting the main findings of the paper.

Section~\ref{sec:token_selection} further studies the role of visual grounding in uncertainty estimation. We show that selecting tokens according to their visual grounding scores provides more informative uncertainty estimates than random token selection (Figure~\ref{fig:token_selection}). We also perform an intervention-based experiment by masking the most visually relevant tokens, showing a stronger effect than random masking and providing additional evidence that the proposed attention-based grounding score identifies visual information that is important for the model's prediction.
Section~\ref{sec:app_Additional_Experiments} presents additional experiments that further analyze the behavior of our method under different settings.
Section~\ref{sec:Results} reports the complete quantitative results for all evaluated models, datasets, and uncertainty estimation methods.
Section~\ref{app_hyperparam} provides the hyperparameters and implementation details used in our experiments to facilitate reproducibility.
Finally, Section~\ref{sec:architecture} analyzes how visual information influences the hidden representations of the model across different layers.

Overall, these additional analyses further support our main finding that \textbf{visual grounding provides useful information for identifying which generated tokens should contribute most to uncertainty estimation}.

\section{Experimental protocol}
\label{sec:Experimental_Protocol}

\subsection{Datasets}
We evaluate VIG-TUQ on several Visual Question Answering (VQA) datasets: \textbf{ADVQA}~\citep{li2021adversarial}, which consists of a training set of $5,000$ samples and a test set of $2,000$ samples, both drawn from the original training set; 
\textbf{VQARAD}~\citep{lau2018dataset}, which includes $2{,}000$ samples from the original training set and $500$ samples from the original test set;
\textbf{OKVQA}~\citep{marino2019ok}, which contains $5{,}000$ training samples drawn from the original training set and $2{,}000$ test samples from the validation set;
and \textbf{VizWiz}~\citep{gurari2018vizwiz}, which contains $5{,}000$ training samples and $2{,}000$ test samples drawn from the original training set.

\subsection{Models}

We evaluate VIG-TUQ on widely used Large Vision Language Models (LVLMs). 
In particular, we conduct experiments on \texttt{Qwen2.5-VL-3B} and \texttt{Qwen2.5-VL-7B}~\citep{bai2025qwen2}, \texttt{Qwen3.5-2B}~\citep{qwen3.5}, \texttt{LLava1.5-7B}, \texttt{LLava1.5-13B}~\citep{liu2024improved}, \texttt{idefics-9B}~\citep{laurenccon2023obelics}  and \texttt{idefics2-8B}~\citep{laurenccon2023obelics}. 
We include models with different architectures to cover the range of modality fusion strategies discussed in Section~\ref{sec:modality_fusion} and to analyze how these fusion strategies affect the visual grounding.
In particular, \texttt{llava}, \texttt{idefics2} and \texttt{Qwen-VL} models adopt a \textit{early fusion} approach, while \texttt{idefics-9B} model rely on an \textit{late fusion} approach and \texttt{Qwen3.5} on a \textit{native} fusion one.

\subsection{Metrics}
\label{app_metrics}

\paragraph{AUROC.}
We assign a probability score to each example for belonging to the positive class (positive reflects that the LVLM is certain about its answer). 
Area Under the ROC Curve (AUROC)~\citep{hendrycks2016baseline} measures how well these scores rank true positives above true negatives. 
It is the chance that a randomly chosen positive example receives a higher score than a randomly chosen negative one. 
Higher AUROC means better class separation.

\paragraph{ECE.}
Expected Calibration Error (ECE)~\citep{guo2017calibration} measures how well a model's predicted confidence aligns with its empirical accuracy. A low ECE indicates that the model is well calibrated, meaning that predictions made with confidence $c$ are correct approximately $c$ fraction of the time. 
Conversely, a high ECE indicates miscalibration, for example when the model is systematically overconfident or underconfident. Since ECE is defined for confidence scores in $[0,1]$, we first normalize all UQ scores to this range before computing ECE.

\section{Token selection analysis}
\label{sec:token_selection}

In addition to Section~\ref{sub_sec:visual_grounding}, we conducted additional experiments to evaluate whether visual grounding scores help identify the tokens most relevant for uncertainty estimation; we conducted a token selection experiment. 

\paragraph{Effect of grounding-based token selection.}
For each generated answer $\vy=[\vy_1,\dots,\vy_T]$, we compute the entropy of each token, $\mathcal{H}(\vy_t \mid \vx)$ (\eqref{def:token_entropy}).
We then rank the tokens using either the distribution-based grounding score $S_{\mathrm{JSD}}(\vy_t,\vx)$ (\eqref{def:js_raw_score}) or the attention-based grounding score $S_A^{(l)}(\vy_t,\vx)$ (\eqref{def:S_A}).
For each ranking, we keep only the top $k\%$ of tokens and compute the final uncertainty score as the sum of the entropies of the selected tokens.
As a baseline, we repeat the same experiment by randomly selecting $k\%$ of the generated tokens.
To reduce the effect of randomness, this random selection is repeated $10$ times for each model, dataset, and value of $k$, and we report the average AUROC.
Figure~\ref{fig:token_selection} shows the results for different LVLM architectures.

\paragraph{Intervention-based validation of the visual grounding score.}
Attention weights do not provide a complete causal explanation of how much the model relies on the image.
For this reason, in VIG-TUQ, the attention-based grounding score is not used alone as an uncertainty score or as a saliency map.
Instead, it is used only to weight token-level entropy and is combined with the complementary distribution-based JSD score.

To further evaluate whether the attention-based score identifies visual regions that are important for the model prediction, we perform an intervention experiment on VQA-RAD with Qwen2.5-VL-3B.
For each generated token, we compare three settings:
\begin{enumerate}
    \item \textbf{Original image:} the token probability is obtained using the unchanged input image.
    \item \textbf{Attention-guided masking:} we mask the $10\%$ of visual tokens with the highest attention-based grounding scores and recompute the token probability.
    \item \textbf{Random masking:} we mask the same number of visual tokens, but select them uniformly at random.
\end{enumerate}

For each generated answer, we average the token-level probabilities over the whole sequence to obtain a sentence-level score.
These scores are then used to predict whether the answer is correct, and performance is evaluated using AUROC.

\begin{table}[t]
  \centering
  \caption{Effect of visual-token masking on the ability of the mean token
  probability to predict answer correctness (AUROC) on VQA-RAD with
  Qwen2.5-VL-3B. Both interventions remove the same amount of visual context.}
  \label{tab:masking-intervention}
  \resizebox{0.5\textwidth}{!}{
  \begin{tabular}{lc}
    \toprule
    Score & AUROC \\
    \midrule
    Mean token probability (unperturbed image) & $0.6872$ \\
    Mean token probability (random masking)    & $0.6294$ \\
    Mean token probability (attention-guided masking) & $0.5688$ \\
    \bottomrule
  \end{tabular}
  }
\end{table}
\joseph{
As reported in Table~\ref{tab:masking-intervention}, masking the visual tokens
selected by our grounding score degrades AUROC substantially more than masking
an equal number of randomly chosen visual tokens ($0.5688$ versus $0.6294$,
against $0.6872$ without perturbation). Since both interventions remove the
same quantity of visual context, this gap indicates that the grounding score
identifies visual information that is particularly influential for the model's
prediction. We stress that this experiment does not constitute a pixel-level
causal explanation, and that masking may itself induce a distribution shift;
it nonetheless provides considerably stronger evidence than a purely
correlational analysis.}

\begin{figure*}[t!]
    \centering
    \begin{center}
        \includegraphics[width=0.9\linewidth]{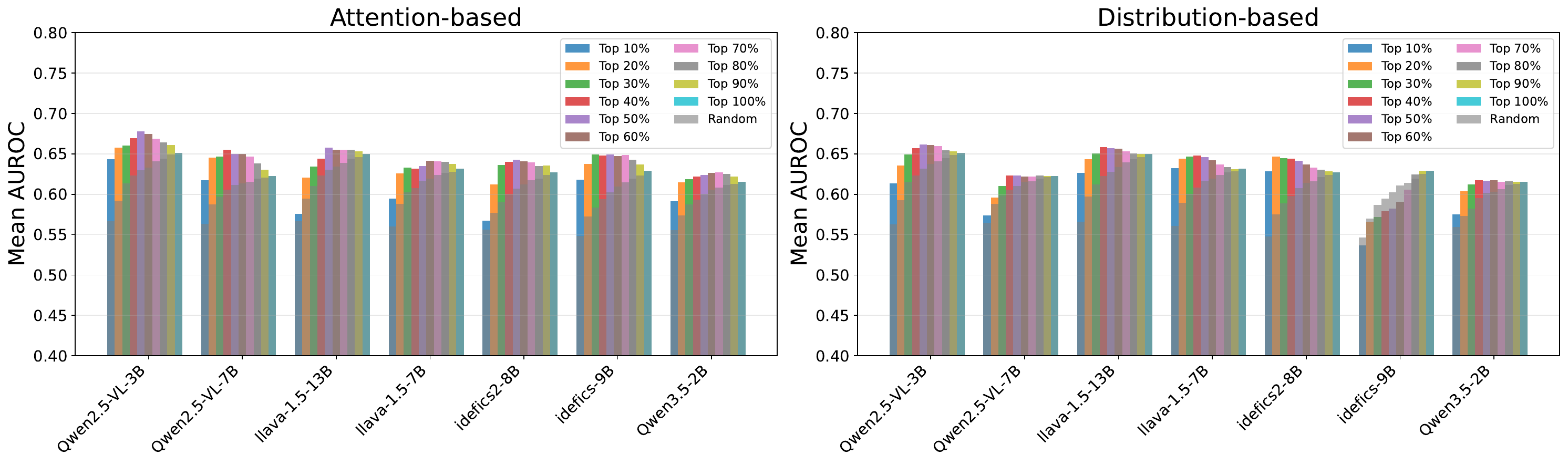}
    \end{center}
    \caption{\textbf{Visual grounding scores help identify the tokens most relevant for uncertainty estimation.}
    AUROC performance reported for the sum of top k\% token entropies across different LVLM architectures. 
    Top k\%  token entropies are selected according to either the attention-based score the the distribution-based score (\eqref{def:js_raw_score}), attention-based score (\eqref{def:S_A}), or randomly from generated tokens. 
    Gray bars correspond to a random token-selection baseline and averaged over $10$ runs.  
    All results are averaged across all datasets.}
    \label{fig:token_selection}
\end{figure*}

\section{Additional experiments}
\label{sec:app_Additional_Experiments}

\paragraph{\textbf{Impact of LVLM architecture.}}

\begin{figure*}[t!]
    \centering
    \begin{center}
        \includegraphics[width=0.8\linewidth]{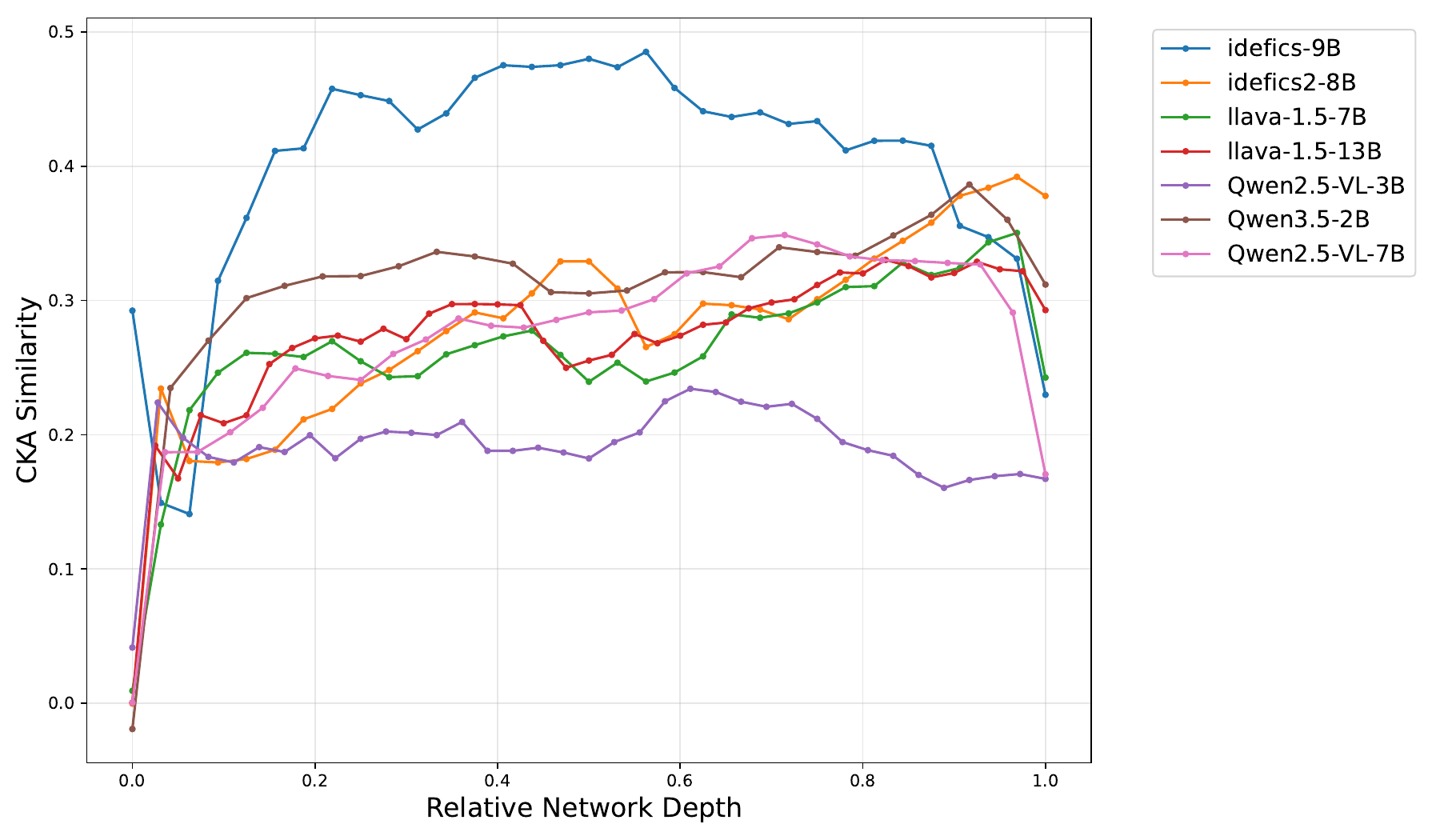}
    \end{center}
    \caption{CKA similarity between LVLM hidden states and reference vision features across normalized network depth. 
For each model, we compare layer-wise hidden representations to the final-layer representation of a frozen vision encoder. 
Higher CKA values indicate stronger alignment with the original visual representation. The late-fusion model \texttt{idefics-9B} exhibits the highest visual alignment across much of the network, consistent with architectures that continue to inject visual information throughout the transformer layers.
}
    \label{fig:CKA}
\end{figure*}
Figure~\ref{fig:CKA} aims to analyze how different LVLM architectures use visual information across their layers. 
To do so, we compute the Centered Kernel Alignment (CKA) similarity between the hidden representations of each LVLM (on \textbf{OKVQA}) and the final-layer features of a reference vision encoder, \texttt{DINOv2} \citep{oquab2023dinov2}. We use \texttt{DINOv2} as a reference because it is trained for visual representation learning and captures strong image features across many vision tasks. 
As a result, it provides a useful proxy for purely visual information. 
We therefore measure how similar LVLM representations are to these visual features.
For each LVLM, we extract hidden states across all transformer layers and compare them to the reference visual features. 
Because the evaluated models have different numbers of layers, we normalize layer indices to a relative depth scale from $0.0$, corresponding to the embedding layer, to $1.0$, corresponding to the final transformer layer. 
This enables a direct comparison of where visual information is most strongly reflected across architectures.

The resulting plot shows how closely each model's internal representations align with the original visual features as generation progresses through the network. 
High CKA similarity indicates that the model representation remains strongly related to the visual encoder representation, whereas lower similarity suggests a shift toward language- or task-specific representations. 
We observe that \texttt{idefics-9B}, the only late-fusion architecture in this comparison, exhibits the strongest similarity to the reference vision encoder. 
This behavior is consistent with its architectural design: unlike early-fusion models, which inject visual features mainly before the language-model layers, late-fusion models continue to integrate visual components throughout the network, leading to stronger visual alignment at intermediate or deeper layers.

\subsection{Computational cost}

We assess the efficiency of \method~ by comparing its average inference time against existing token-level UQ methods on OKVQA. Table ~\ref{tab:speed_test} reports the mean runtime for answer generation and uncertainty estimation, measured with the same backbone models and hardware. Because every token-level method follows an identical decoding procedure, generation time is fixed across methods, so all runtime gaps stem solely from the post-hoc UQ computation. CCP and Token-SAR rely on external models and therefore incur extra computational cost, whereas \method~ avoids this by reusing only the token probabilities and attention weights already produced by the generation model. Lightweight baselines such as Token-Entropy, Log-Perplexity, and Max-Prob are still marginally faster, but \method~ adds only negligible overhead and is considerably more efficient than the remaining token-level UQ approaches.

\begin{table}[!t]
\centering
\small
\setlength{\tabcolsep}{4pt}
\caption{\textbf{VIG-TUQ is computationally efficient.}
This table shows the average inference time for: answer generation and uncertainty quantification on the \textit{OKVQA} dataset.}
\label{tab:speed_test}
\resizebox{1.0\columnwidth}{!}{
\begin{tabular}{c|ccccc|c}
\toprule
\textbf{Model}
& \textit{Max-Prob}
& \textit{Token-Entropy}
& \textit{CCP}
& \textit{Token-SAR}
& \cellcolor{green!20}\textit{\method}
& \textit{KLE} \\
\midrule
\texttt{Qwen2.5-3B}
& 1.01s & 1.00s & 3.26s & 1.71s & \cellcolor{green!20}1.29s & 4.78s \\
\texttt{Qwen2.5-7B}
& 1.58s & 1.60s & 5.28s & 2.39s & \cellcolor{green!20}2.00s & 8.42s \\
\texttt{llava-1.5-7B}
& 0.81s & 0.84s & 2.81s & 1.09s & \cellcolor{green!20}1.01s & 4.83s \\
\texttt{llava-1.5-13B}
& 1.39s & 1.41s & 4.64s & 2.10s & \cellcolor{green!20}1.75s & 7.40s \\
\texttt{idefics2-8B}
& 0.65s & 0.66s & 2.18s & 0.99s & \cellcolor{green!20}0.82s & 3.47s \\
\texttt{idefics-9B}
& 0.72s & 0.68s & 2.30s & 1.06s & \cellcolor{green!20}1.20s & 4.60s \\
\texttt{Qwen3.5-2B}
& 0.98s & 0.88s & 2.98s & 1.58s & \cellcolor{green!20}1.52s & 6.26s \\
\midrule
\textbf{Mean}
& 1.02s & 1.01s & 3.35s & 1.56s & \cellcolor{green!20}1.37s & 5.68s \\
\bottomrule
\end{tabular}
}
\end{table}

\section{Complete results}
\label{sec:Results}

In this section, we present the detailed performance of all UQ methods across datasets and models described in Appendix~\ref{sec:Experimental_Protocol}. 
Table~\ref{tab:fuull_results} reports the AUROC scores achieved by each method on every dataset and LVLM architecture.

\paragraph{Comparision with token-level UQ methods.}
VIG-TUQ consistently outperforms conventional token-level uncertainty metrics, including Log-Perplexity, Max-Prob, Token-Entropy, CCP, and Token-SAR.
For instance, on the \textbf{OKVQA} dataset, \method~ achieves the highest AUROC score. 
Similar trends are observed for \textbf{VQARAD} and \textbf{ADVQA}, where VIG-TUQ ranks among the top-performing methods.
These results confirm that integrating multiple importance scores and visual grounding scores improves the performance of token-level UQ methods.

\paragraph{Comparison with consistency-based methods.}
Semantic Entropy (SE)~\citep{farquhar2024detecting} and Kernel Language Entropy (KLE)~\citep{nikitin2024kernel} achieve slightly higher AUROC on certain datasets. 
This behavior is expected as these methods estimate uncertainty by measuring the semantic consistency of generated responses, which provides a more comprehensive assessment of uncertainty across the entire generated content. 
However, these methods require multiple sampling, which also makes them significantly more expensive at inference time (see Table~\ref{tab:speed_test}). 
In contrast, our approach requires only a single or two samplings and therefore remains much more efficient while still maintaining competitive performance.

\paragraph{Effect across datasets.}
We also observe that the performance of UQ methods varies across datasets. 
For instance, \method~ clearly outperforms on \textbf{OKVQA} and \textbf{VQARAD}, which require stronger reasoning and better exploitation of the visual content, which makes reliable uncertainty estimation more challenging. 
On \textbf{VizWiz}, the improvements are still visible but slightly smaller. 
This dataset contains real-world images with visual noise, which can make the visual grounding more difficult for all models.

\paragraph{Effect across model architectures.}
The benefits of VIG-TUQ are consistent across different LVLM architectures. 
This suggests that VIG-TUQ is not tied to a specific architecture and can be effectively applied across a wide range of multimodal architectures.

\begin{table*}[]
\caption{\textbf{\method ~outperforms token-level UQ baselines.} Evaluation metrics for LVLM architectures across all datasets and models.}
\centering
\label{tab:fuull_results}
\resizebox{1.0\textwidth}{!}{
\begin{tabular}{cc|cccccccccccccc}
\toprule
 &  & \texttt{Qwen2.5-VL-3B} & \texttt{Qwen2.5-VL-3B} & \texttt{Qwen2.5-VL-7B} & \texttt{Qwen2.5-VL-7B} & \texttt{llava-1.5-13B} & \texttt{llava-1.5-13B} & \texttt{llava-1.5-7B} & \texttt{llava-1.5-7B} & \texttt{idefics2-8B} & \texttt{idefics2-8B} & \texttt{idefics-9B} & \texttt{idefics-9B} & \texttt{Qwen3.5-2B} & \texttt{Qwen3.5-2B} \\
\multicolumn{1}{l}{} & \multicolumn{1}{l}{} & AUROC$\uparrow$ & ECE$\downarrow$ & AUROC$\uparrow$ & ECE$\downarrow$ & AUROC$\uparrow$ & ECE$\downarrow$ & AUROC$\uparrow$ & ECE$\downarrow$ & AUROC$\uparrow$ & ECE$\downarrow$ & AUROC$\uparrow$ & ECE$\downarrow$ & AUROC$\uparrow$ & ECE$\downarrow$ \\
\midrule
\textbf{OKVQA} & \textit{Log Perplexity} & 0.610 (0.008) & 0.110 (0.007) & 0.595 (0.008) & 0.109 (0.007) & 0.635 (0.008) & 0.139 (0.007) & 0.634 (0.008) & 0.118 (0.006) & 0.637 (0.008) & 0.146 (0.006) & 0.605 (0.008) & 0.104 (0.007) & 0.649 (0.008) & 0.071 (0.006) \\
\textbf{OKVQA} & \textit{Token-Sar} & 0.551 (0.008) & 0.159 (0.007) & 0.581 (0.008) & 0.115 (0.007) & 0.639 (0.008) & 0.145 (0.007) & 0.637 (0.008) & 0.114 (0.007) & 0.604 (0.008) & 0.159 (0.007) & 0.524 (0.008) & 0.158 (0.007) & 0.600 (0.008) & 0.110 (0.007) \\
\textbf{OKVQA} & \textit{CCP} & 0.607 (0.008) & 0.111 (0.007) & \underline{0.611} (0.008) & \textbf{0.088} (0.007) & 0.614 (0.008) & 0.168 (0.007) & 0.612 (0.008) & 0.138 (0.007) & 0.647 (0.008) & 0.172 (0.006) & \textbf{0.662} (0.008) & \textbf{0.068} (0.006) & 0.625 (0.008) & \textbf{0.043} (0.008) \\
\textbf{OKVQA} & \textit{Max-Prob} & 0.592 (0.008) & 0.153 (0.007) & 0.608 (0.008) & 0.113 (0.007) & 0.611 (0.008) & 0.150 (0.007) & 0.609 (0.008) & 0.152 (0.007) & 0.634 (0.008) & \underline{0.139} (0.006) & 0.644 (0.008) & 0.108 (0.006) & 0.649 (0.008) & 0.070 (0.006) \\
\textbf{OKVQA} & \textit{Token-Entropy} & 0.610 (0.008) & 0.109 (0.007) & 0.601 (0.008) & 0.102 (0.007) & 0.641 (0.008) & 0.139 (0.007) & 0.638 (0.008) & 0.116 (0.006) & 0.642 (0.008) & 0.145 (0.007) & 0.608 (0.008) & 0.098 (0.007) & \underline{0.651} (0.008) & \underline{0.069} (0.006) \\
\textbf{OKVQA} & \textit{\method}-A (ours) & \underline{0.657} (0.008) & 0.077 (0.006) & \textbf{0.620} (0.008) & 0.092 (0.007) & 0.644 (0.008) & \underline{0.136} (0.007) & \underline{0.651} (0.008) & \underline{0.106} (0.006) & 0.664 (0.008) & 0.143 (0.006) & \underline{0.645} (0.008) & \underline{0.072} (0.007) & \textbf{0.652} (0.008) & 0.073 (0.006) \\
\textbf{OKVQA} & \textit{\method}-JSD (ours) & 0.652 (0.008) & \textbf{0.019} (0.007) & 0.606 (0.008) & \underline{0.090} (0.007) & \underline{0.649} (0.008) & \textbf{0.049} (0.006) & 0.646 (0.008) & \textbf{0.065} (0.007) & \underline{0.669} (0.008) & \textbf{0.061} (0.006) & 0.545 (0.008) & 0.140 (0.007) & 0.633 (0.008) & 0.121 (0.006) \\
 \rowcolor{green!20} \textbf{OKVQA} & \textit{\method} (ours) & \textbf{0.664} (0.008) & \underline{0.071} (0.006) & \textbf{0.620} (0.008) & 0.092 (0.007) & \textbf{0.657} (0.008) & 0.137 (0.007) & \textbf{0.657} (0.008) & 0.109 (0.006) & \textbf{0.674} (0.008) & 0.144 (0.007) & \underline{0.645} (0.008) & \underline{0.072} (0.007) & \textbf{0.652} (0.008) & 0.073 (0.007) \\
\textbf{OKVQA} & \textit{SE} & 0.669 (0.008) & 0.041 (0.006) & 0.647 (0.008) & 0.080 (0.007) & 0.690 (0.008) & 0.076 (0.007) & 0.708 (0.008) & 0.075 (0.006) & 0.702 (0.008) & 0.110 (0.006) & 0.706 (0.008) & 0.035 (0.007) & 0.529 (0.008) & 0.099 (0.006) \\
\textbf{OKVQA} & \textit{KLE} & 0.682 (0.008) & 0.035 (0.007) & 0.654 (0.008) & 0.078 (0.007) & 0.700 (0.008) & 0.074 (0.006) & 0.717 (0.008) & 0.069 (0.006) & 0.715 (0.008) & 0.110 (0.007) & 0.721 (0.008) & 0.030 (0.006) & 0.643 (0.008) & 0.112 (0.007) \\
\midrule

\textbf{VIZWIZ} & \textit{Log Perplexity} & 0.690 (0.012) & 0.088 (0.010) & 0.634 (0.013) & 0.131 (0.011) & 0.652 (0.012) & 0.081 (0.009) & 0.667 (0.012) & 0.076 (0.010) & 0.635 (0.012) & 0.108 (0.010) & 0.612 (0.013) & 0.121 (0.010) & 0.634 (0.012) & 0.082 (0.010) \\
\textbf{VIZWIZ} & \textit{Token-Sar} & 0.525 (0.013) & 0.185 (0.011) & 0.560 (0.013) & 0.181 (0.012) & 0.636 (0.012) & 0.082 (0.010) & 0.628 (0.012) & 0.075 (0.010) & 0.597 (0.013) & 0.132 (0.010) & 0.576 (0.013) & 0.126 (0.011) & 0.591 (0.013) & 0.112 (0.010) \\
\textbf{VIZWIZ} & \textit{CCP} & 0.665 (0.012) & 0.121 (0.010) & 0.612 (0.013) & 0.155 (0.011) & 0.639 (0.012) & 0.096 (0.010) & 0.655 (0.013) & 0.060 (0.011) & 0.639 (0.012) & 0.132 (0.011) & \textbf{0.655} (0.012) & \textbf{0.063} (0.011) & 0.615 (0.012) & 0.084 (0.012) \\
\textbf{VIZWIZ} & \textit{Max-Prob} & 0.663 (0.012) & 0.108 (0.010) & 0.614 (0.013) & 0.132 (0.011) & 0.637 (0.012) & 0.136 (0.010) & 0.660 (0.013) & 0.130 (0.010) & 0.632 (0.012) & 0.121 (0.010) & \underline{0.652} (0.012) & 0.129 (0.011) & 0.631 (0.012) & 0.087 (0.011) \\
\textbf{VIZWIZ} & \textit{Token-Entropy} & 0.686 (0.012) & 0.088 (0.010) & 0.637 (0.013) & 0.127 (0.011) & 0.662 (0.012) & \underline{0.080} (0.010) & 0.673 (0.012) & 0.072 (0.010) & 0.639 (0.012) & \underline{0.104} (0.010) & 0.618 (0.013) & 0.123 (0.011) & 0.632 (0.012) & 0.080 (0.010) \\
\textbf{VIZWIZ} & \textit{\method}-A (ours) & \underline{0.722} (0.011) & \textbf{0.085} (0.010) & \underline{0.680} (0.012) & \textbf{0.103} (0.011) & \textbf{0.674} (0.012) & \textbf{0.076} (0.010) & \underline{0.686} (0.012) & \underline{0.059} (0.010) & \underline{0.664} (0.012) & \textbf{0.099} (0.011) & 0.641 (0.013) & \underline{0.113} (0.011) & \underline{0.667} (0.012) & \textbf{0.066} (0.010) \\
\textbf{VIZWIZ} & \textit{\method}-JSD (ours) & 0.721 (0.012) & \underline{0.086} (0.010) & \underline{0.680} (0.012) & 0.112 (0.011) & 0.645 (0.012) & 0.086 (0.010) & 0.679 (0.012) & \textbf{0.058} (0.009) & 0.659 (0.012) & 0.111 (0.011) & 0.603 (0.013) & 0.157 (0.011) & 0.660 (0.012) & \underline{0.071} (0.010) \\
 \rowcolor{green!20}\textbf{VIZWIZ} & \textit{\method} (ours) & \textbf{0.727} (0.011) & \textbf{0.085} (0.010) & \textbf{0.685} (0.012) & \underline{0.105} (0.011) & \underline{0.673} (0.012) & \textbf{0.076} (0.010) & \textbf{0.692} (0.012) & \underline{0.059} (0.010) & \textbf{0.668} (0.012) & 0.110 (0.011) & 0.641 (0.013) & \underline{0.113} (0.011) & \textbf{0.669} (0.012) & \textbf{0.066} (0.010) \\
\textbf{VIZWIZ} & \textit{SE} & 0.694 (0.012) & 0.035 (0.010) & 0.675 (0.012) & 0.118 (0.011) & 0.657 (0.012) & 0.039 (0.010) & 0.676 (0.012) & 0.020 (0.010) & 0.644 (0.012) & 0.067 (0.010) & 0.673 (0.012) & 0.062 (0.011) & 0.521 (0.013) & 0.084 (0.010) \\
\textbf{VIZWIZ} & \textit{KLE} & 0.716 (0.012) & 0.036 (0.010) & 0.687 (0.012) & 0.118 (0.011) & 0.689 (0.012) & 0.042 (0.010) & 0.699 (0.012) & 0.028 (0.010) & 0.672 (0.012) & 0.066 (0.010) & 0.699 (0.012) & 0.065 (0.011) & 0.614 (0.013) & 0.089 (0.010) \\
\midrule

\textbf{VQARAD} & \textit{Log Perplexity} & 0.666 (0.027) & 0.176 (0.021) & 0.646 (0.030) & 0.255 (0.019) & 0.660 (0.028) & 0.217 (0.021) & \underline{0.604} (0.029) & 0.219 (0.021) & 0.599 (0.027) & 0.179 (0.020) & 0.687 (0.025) & 0.157 (0.022) & 0.612 (0.026) & 0.128 (0.021) \\
\textbf{VQARAD} & \textit{Token-Sar} & 0.611 (0.027) & 0.180 (0.020) & 0.637 (0.030) & 0.254 (0.019) & 0.656 (0.028) & 0.216 (0.022) & 0.589 (0.030) & 0.221 (0.022) & 0.599 (0.027) & 0.173 (0.022) & 0.595 (0.027) & 0.162 (0.021) & 0.628 (0.026) & 0.119 (0.021) \\
\textbf{VQARAD} & \textit{CCP} & 0.661 (0.027) & \textbf{0.148} (0.022) & \underline{0.653} (0.030) & \underline{0.233} (0.020) & 0.608 (0.030) & \textbf{0.196} (0.022) & 0.595 (0.030) & \textbf{0.185} (0.022) & 0.587 (0.027) & \textbf{0.127} (0.020) & 0.657 (0.026) & \textbf{0.148} (0.020) & 0.589 (0.027) & 0.144 (0.022) \\
\textbf{VQARAD} & \textit{Max-Prob} & \underline{0.674} (0.026) & \underline{0.164} (0.021) & 0.649 (0.029) & 0.252 (0.019) & 0.618 (0.030) & 0.232 (0.020) & 0.602 (0.029) & 0.222 (0.019) & 0.568 (0.028) & 0.202 (0.022) & 0.664 (0.026) & 0.192 (0.022) & 0.611 (0.026) & 0.128 (0.022) \\
\textbf{VQARAD} & \textit{Token-Entropy} & 0.669 (0.027) & 0.171 (0.021) & 0.639 (0.031) & 0.258 (0.019) & 0.659 (0.028) & 0.225 (0.020) & 0.603 (0.030) & 0.217 (0.021) & 0.605 (0.027) & 0.177 (0.021) & \underline{0.696} (0.025) & \underline{0.154} (0.022) & 0.624 (0.026) & 0.123 (0.022) \\
\textbf{VQARAD} & \textit{\method}-A (ours) & \textbf{0.712} (0.026) & 0.168 (0.021) & \textbf{0.676} (0.030) & 0.256 (0.019) & 0.672 (0.028) & 0.260 (0.021) & 0.598 (0.029) & 0.253 (0.021) & \underline{0.665} (0.027) & \underline{0.147} (0.021) & \textbf{0.736} (0.024) & \underline{0.154} (0.022) & \underline{0.634} (0.026) & 0.117 (0.021) \\
\textbf{VQARAD} & \textit{\method}-JSD (ours) & 0.616 (0.027) & 0.194 (0.021) & 0.540 (0.030) & \textbf{0.205} (0.019) & \underline{0.689} (0.028) & 0.245 (0.021) & \textbf{0.670} (0.029) & 0.230 (0.021) & \textbf{0.683} (0.027) & 0.246 (0.021) & 0.636 (0.026) & 0.280 (0.022) & 0.631 (0.026) & \textbf{0.069} (0.021) \\
 \rowcolor{green!20}\textbf{VQARAD} & \textit{\method} (ours) & \textbf{0.712} (0.026) & 0.168 (0.021) & \textbf{0.676} (0.030) & 0.256 (0.019) & \textbf{0.693} (0.028) & \underline{0.215} (0.021) & \textbf{0.670} (0.028) & \underline{0.201} (0.021) & 0.658 (0.027) & 0.149 (0.022) & \textbf{0.736} (0.024) & \underline{0.154} (0.022) & \textbf{0.641} (0.026) & \underline{0.116} (0.020) \\
\textbf{VQARAD} & \textit{SE} & 0.685 (0.026) & 0.168 (0.021) & 0.632 (0.030) & 0.197 (0.019) & 0.630 (0.029) & 0.258 (0.021) & 0.683 (0.029) & 0.246 (0.021) & 0.611 (0.027) & 0.125 (0.021) & 0.669 (0.026) & 0.145 (0.021) & 0.651 (0.026) & 0.082 (0.021) \\
\textbf{VQARAD} & \textit{KLE} & 0.679 (0.026) & 0.169 (0.021) & 0.630 (0.030) & 0.198 (0.019) & 0.644 (0.029) & 0.259 (0.021) & 0.678 (0.029) & 0.247 (0.021) & 0.620 (0.027) & 0.126 (0.021) & 0.666 (0.026) & 0.144 (0.021) & 0.679 (0.026) & 0.078 (0.021) \\
\midrule
\textbf{ADVQA} & \textit{Log Perplexity} & \underline{0.646} (0.015) & 0.143 (0.013) & 0.614 (0.015) & 0.168 (0.012) & 0.638 (0.014) & 0.128 (0.012) & 0.617 (0.015) & 0.138 (0.012) & 0.624 (0.015) & 0.171 (0.012) & 0.599 (0.016) & 0.253 (0.011) & 0.565 (0.015) & 0.174 (0.012) \\
\textbf{ADVQA} & \textit{Token-Sar} & 0.531 (0.016) & 0.194 (0.013) & 0.574 (0.015) & 0.186 (0.012) & 0.629 (0.014) & \textbf{0.105} (0.011) & 0.581 (0.015) & 0.140 (0.011) & 0.598 (0.015) & \underline{0.157} (0.011) & 0.506 (0.015) & 0.249 (0.011) & 0.531 (0.015) & 0.199 (0.011) \\
\textbf{ADVQA} & \textit{CCP} & 0.642 (0.015) & \underline{0.112} (0.012) & 0.603 (0.015) & \textbf{0.119} (0.012) & 0.649 (0.014) & \textbf{0.091} (0.012) & \textbf{0.637} (0.014) & \textbf{0.103} (0.011) & 0.634 (0.015) & \textbf{0.135} (0.012) & \underline{0.663} (0.015) & \underline{0.208} (0.011) & \textbf{0.588} (0.015) & \textbf{0.130} (0.012) \\
\textbf{ADVQA} & \textit{Max-Prob} & 0.642 (0.015) & 0.185 (0.013) & 0.594 (0.015) & 0.197 (0.012) & 0.636 (0.014) & 0.142 (0.013) & 0.624 (0.015) & 0.146 (0.012) & 0.633 (0.015) & 0.194 (0.013) & \textbf{0.665} (0.015) & 0.237 (0.011) & 0.554 (0.015) & 0.172 (0.012) \\
\textbf{ADVQA} & \textit{Token-Entropy} & 0.642 (0.015) & 0.148 (0.012) & 0.612 (0.015) & 0.175 (0.012) & 0.639 (0.014) & 0.128 (0.012) & 0.613 (0.015) & 0.143 (0.012) & 0.624 (0.015) & 0.176 (0.012) & 0.594 (0.016) & 0.250 (0.010) & 0.556 (0.015) & 0.179 (0.011) \\
\textbf{ADVQA} & \textit{\method}-A (ours) & \textbf{0.665} (0.014) & \underline{0.133} (0.013) & \underline{0.626} (0.015) & 0.163 (0.012) & \underline{0.651} (0.014) & 0.116 (0.012) & 0.624 (0.015) & \underline{0.137} (0.012) & \textbf{0.644} (0.015) & 0.167 (0.012) & 0.617 (0.016) & 0.237 (0.011) & \underline{0.577} (0.014) & \underline{0.171} (0.012) \\
\textbf{ADVQA} & \textit{\method}-JSD (ours) & 0.620 (0.015) & 0.150 (0.012) & 0.617 (0.015) & \underline{0.160} (0.012) & 0.643 (0.015) & \underline{0.113} (0.012) & 0.613 (0.015) & 0.142 (0.012) & 0.594 (0.015) & 0.182 (0.012) & 0.500 (0.016) & \textbf{0.213} (0.011) & 0.531 (0.015) & 0.189 (0.012) \\
 \rowcolor{green!20}\textbf{ADVQA} & \textit{\method} (ours) & 0.662 (0.014) & \textbf{0.131} (0.012) & \textbf{0.632} (0.015) & 0.159 (0.012) & \textbf{0.662} (0.014) & 0.112 (0.012) & \underline{0.633} (0.015) & \textbf{0.133} (0.012) & \underline{0.642} (0.015) & 0.165 (0.012) & 0.617 (0.016) & 0.237 (0.011) & 0.567 (0.014) & 0.175 (0.012) \\
\textbf{ADVQA} & \textit{SE} & 0.601 (0.015) & 0.214 (0.012) & 0.651 (0.015) & 0.195 (0.012) & 0.695 (0.014) & 0.135 (0.012) & 0.652 (0.015) & 0.141 (0.012) & 0.662 (0.015) & 0.197 (0.012) & 0.650 (0.015) & 0.243 (0.011) & 0.517 (0.015) & 0.229 (0.012) \\
\textbf{ADVQA} & \textit{KLE} & 0.614 (0.015) & 0.213 (0.012) & 0.660 (0.015) & 0.194 (0.012) & 0.701 (0.014) & 0.132 (0.012) & 0.651 (0.015) & 0.139 (0.012) & 0.666 (0.015) & 0.196 (0.012) & 0.662 (0.015) & 0.242 (0.011) & 0.568 (0.015) & 0.230 (0.012) \\

\midrule
\end{tabular}

}

\end{table*}

\section{Hyperparameters}
\label{app_hyperparam}

The hyperparameters used in \method~ are reported in Table~\ref{tab:hypperparam}. 
They are selected on a small training split for each dataset--model combination. 
To limit the search space and avoid overfitting to the validation setting, we impose a strict constraint on the weighting coefficients: both $\alpha_{\mathrm{JSD}}$ and $\alpha_A$ are restricted to integer values in $[0,5]$. 
This deliberately coarse search is intended to ensure that the performance gains of \method~ do not rely on extensive hyperparameter tuning.
The selected hyperparameters show that some model- and dataset-specific adaptation is beneficial. 
This is consistent with the token selection analysis in Figure~\ref{fig:token_selection}, where the attention-based and distribution-based visual grounding scores do not perform best in the same settings. 
For example, for \texttt{llava-1.5-7B} on \textbf{VQARAD}, \method~ relies only on the distribution-based score, whereas for \texttt{Qwen3.5-2B} on \textbf{OKVQA}, it relies only on the attention-based score. 
Nevertheless, in most model--dataset combinations, both scores receive nonzero weights, suggesting that the two grounding criteria provide complementary information for uncertainty quantification.

\paragraph{\textbf{Hyperparameter sensitivity and tuning requirements.}}
As discussed above, \method~ requires only a lightweight hyperparameter search. 
To avoid relying on extensive tuning, we restrict the weighting coefficients $\alpha_{\mathrm{JSD}}$ and $\alpha_A$ to integer values in $\{0,\dots,5\}$, and tune them together with the attention layer $l$ on a small training split. 
Table~\ref{tab:_training_hp} reports the selected hyperparameters when varying the number of training examples $K$. 
We observe that stable and competitive hyperparameters can already be obtained with as few as $K=100$ examples. 
In several cases, such as Idefics-9B on OKVQA and VQARAD, the selected configuration remains identical across different values of $K$. 
Even when the selected coefficients vary, as for \texttt{Qwen2.5-VL-3B} on OKVQA, the resulting AUROC remains nearly unchanged. 
These results suggest that \method~ is not overly sensitive to the exact hyperparameter choice and does not require large validation sets or fine-grained tuning to achieve strong performance.

\begin{table}[]
\caption{
\textbf{Hyperparameter tuning stability under limited training data. }
For each model--dataset pair, we tune the discrete weighting coefficients $\alpha_{\mathrm{JSD}}, \alpha_A \in \{0,\dots,5\}$ and the attention layer $l$ using only $K$ training examples. The selected hyperparameters and resulting AUROC/ECE remain stable even for small values of $K$, indicating that \method requires only lightweight tuning and is robust to small hyperparameter variations.
}
\label{tab:_training_hp}
\centering
\resizebox{0.7\columnwidth}{!}{
\begin{tabular}{cc|cccccc}
\toprule
&Dataset & K & AUROC$\uparrow$  & ECE$\downarrow$ &  $\alpha_{\mathrm{JSD}}$ & $\alpha_A$ & $l$ \\
 \midrule
\multirow{8}{*}{\rotatebox{90}{\texttt{idefics-9B}}}
&\textbf{OKVQA} & 1000 & 0,645 & 0.072 & 0 & 1 & 16 \\
&\textbf{OKVQA} & 500 & 0,645 & 0.072 & 0 & 1 & 16 \\
&\textbf{OKVQA} & 200 & 0,645 & 0.072 & 0 & 1 & 16 \\
&\textbf{OKVQA} & 100 & 0,645 & 0.073 & 0 & 1 & 16 \\
 &&&&&&&\\
&\textbf{VQARAD} & 1000 & 0.736 & 0.154 & 0 & 1 & 17 \\
&\textbf{VQARAD} & 500 & 0.730 & 0.164 & 1 & 4 & 16 \\
&\textbf{VQARAD} & 200 & 0.736 & 0.168 & 0 & 1 & 17 \\
&\textbf{VQARAD} & 100 & 0.736 & 0.180 & 0 & 1 & 17 \\
 \midrule
 & \multicolumn{1}{l}{} & \multicolumn{1}{l}{} & \multicolumn{1}{l}{} & \multicolumn{1}{l}{} & \multicolumn{1}{l}{} & \multicolumn{1}{l}{} \\
  \midrule
\multirow{8}{*}{\rotatebox{90}{{\texttt{Qwen2.5-VL-3B}}}}
&\textbf{OKVQA} & 1000 & 0.664 & 0.071 & 1 & 4 & 7 \\
&\textbf{OKVQA} & 500 & 0.664 & 0.071 & 3 & 4 & 7 \\
&\textbf{OKVQA} & 200 & 0.664 & 0.078 & 1 & 3 & 7 \\
&\textbf{OKVQA} & 100 & 0.664 & 0.074 & 3 & 4 & 7 \\
 &&&&&&&\\
&\textbf{VQARAD} & 1000 & 0.712 & 0.168 & 0 & 1 & 6 \\
&\textbf{VQARAD} & 500 & 0.712 & 0.168 & 0 & 1 & 6 \\
&\textbf{VQARAD} & 200 & 0.712 & 0.168 & 0 & 1 & 6 \\
&\textbf{VQARAD} & 100 & 0.712 & 0.168& 0 & 1 & 6\\
\midrule
\end{tabular}
}
\end{table}

\begin{table*}[]
\caption{}
\label{tab:hypperparam}
\resizebox{1.0\textwidth}{!}{
\begin{tabular}{c|ccccccccccccccccccccc}
\toprule
\textbf{} & \multicolumn{3}{c}\texttt{Qwen2.5-VL-3B} & \multicolumn{3}{c}\texttt{Qwen2.5-VL-7B} & \multicolumn{3}{c}\texttt{llava-1.5-13B} & \multicolumn{3}{c}\texttt{llava-1.5-7B} & \multicolumn{3}{c}\texttt{idefics2-8B} & \multicolumn{3}{c}\texttt{idefics-9B} & \multicolumn{3}{c}\texttt{Qwen3.5-2B} \\
\midrule
Dataset &  $\alpha_{\mathrm{JSD}}$ & $\alpha_A$ & $l$ & $\alpha_{\mathrm{JSD}}$ & $\alpha_A$ & $l$ & $\alpha_{\mathrm{JSD}}$ & $\alpha_A$ & $l$ & $\alpha_{\mathrm{JSD}}$ & $\alpha_A$ & $l$ & $\alpha_{\mathrm{JSD}}$ & $\alpha_A$ & $l$ & $\alpha_{\mathrm{JSD}}$ & $\alpha_A$ & $l$ & $\alpha_{\mathrm{JSD}}$ & $\alpha_A$ & $l$ \\
\midrule
\textbf{OKVQA} & 1 & 4 & 7 & 0 & 1 & 4 & 2 & 3 & 6 & 2 & 3 & 16 & 1 & 4 & 29 & 0 & 1 & 16 & 0 & 1 & 5 \\
\textbf{VIZWIZ} & 3 & 1 & 9 & 2 & 3 & 15 & 1 & 4 & 16 & 3 & 4 & 8 & 1 & 1 & 20 & 0 & 1 & 14 & 1 & 3 & 5 \\
\textbf{VQARAD} & 0 & 1 & 6 & 0 & 1 & 6 & 3 & 1 & 3 & 1 & 0 & 18 & 3 & 2 & 0 & 0 & 1 & 17 & 1 & 1 & 5 \\
\textbf{ADVQA} & 1 & 4 & 7 & 1 & 3 & 4 & 1 & 4 & 6 & 1 & 3 & 4 & 1 & 4 & 1 & 0 & 1 & 14 & 1 & 3 & 0
\end{tabular}
}

\end{table*}

\section{Analysis of visual influence on hidden representations}
\label{sec:architecture}

In this section, we examine how the internal representations of an LVLM change when visual input is removed. 
In particular, we compare the hidden representations obtained during normal generation with those obtained when the image is replaced by a blank input.

\paragraph{LVLM.}
Let $\vx=(\vx_v,\vx_l)$ be a multimodal input with a visual content $\vx_v$ and a textual content $\vx_l$. 
As described in Section~\ref{sec:preliminaries}, an LVLM defines a parameterized function $f_\vtheta: \mathcal{X}_v \times \mathcal{X}_l \to \R^d$ to encode the visual-linguistic context into a latent representation.
In particular, the parametric function $f_\vtheta: \mathcal{X}_v \times \mathcal{X}_l \to \R^d$ can be defined recursively over $L$ layers as
\begin{align*}
     f_{\vtheta}(\vx_v,\vx_l) &= f_{L}(\vh_{L-1},\vx_v), \qquad \text{where} \\
    \vh_i &= f_{i}(\vh_{i-1},\vx_v) = g_i\bigl(m_{i}\bigl(E_i(\vx_v), \vh_{i-1}\bigr)\bigr) \qquad, \\ (1 \leq i \leq L) \\
    \vh_0 &=  E_l(\vx_{l});
\end{align*}
with  $E_l(\cdot)$ a text embedding function. 
For each layer $i$, $g_i: \R^d \to \R^d$ depicts a transformer decoder layer, $m_{i}(\cdot, \cdot)$ denotes a multimodal fusion module (e.g., concatenation, linear layer, cross-attention layer), and $E_i(\cdot)$ is a visual encoder (e.g., ViT, CNNs, identity function).

\paragraph{Visual influence analysis problem.}
Our goal is to measure how much the hidden representations change at each layer $l$ when the visual component $\vx_v$ is removed. 
To this end, we examine the internal representations of the LVLM that rely only on the textual component $\vx_l$ defined as
\begin{align}
    \vh_i' &= f_{i}(\vh_{i-1},\varnothing) = g_i\bigl(m_{i}\bigl(\varnothing, \vh_{i-1}'\bigr)\bigr) \qquad (1 \leq i \leq L) \\
    \vh_0' &=  E_l(\vx_{l});
\end{align}
To quantify the difference between these two embeddings at each layer $i$, we compute the cosine distance between $\vh_i$ and $\vh_i'$ at each layer. 
This distance reflects how much the model relies on visual information, where a larger value means that the image exerts a stronger influence on the internal representation.
For each generated answer, we extract the hidden state corresponding to the last token at every layer. 
We then compute the cosine distance between the two representations and average the results across all samples of the dataset.

\begin{figure*}[t]
\centering

\begin{subfigure}{0.32
\linewidth}
    \centering
    \includegraphics[width=\linewidth]{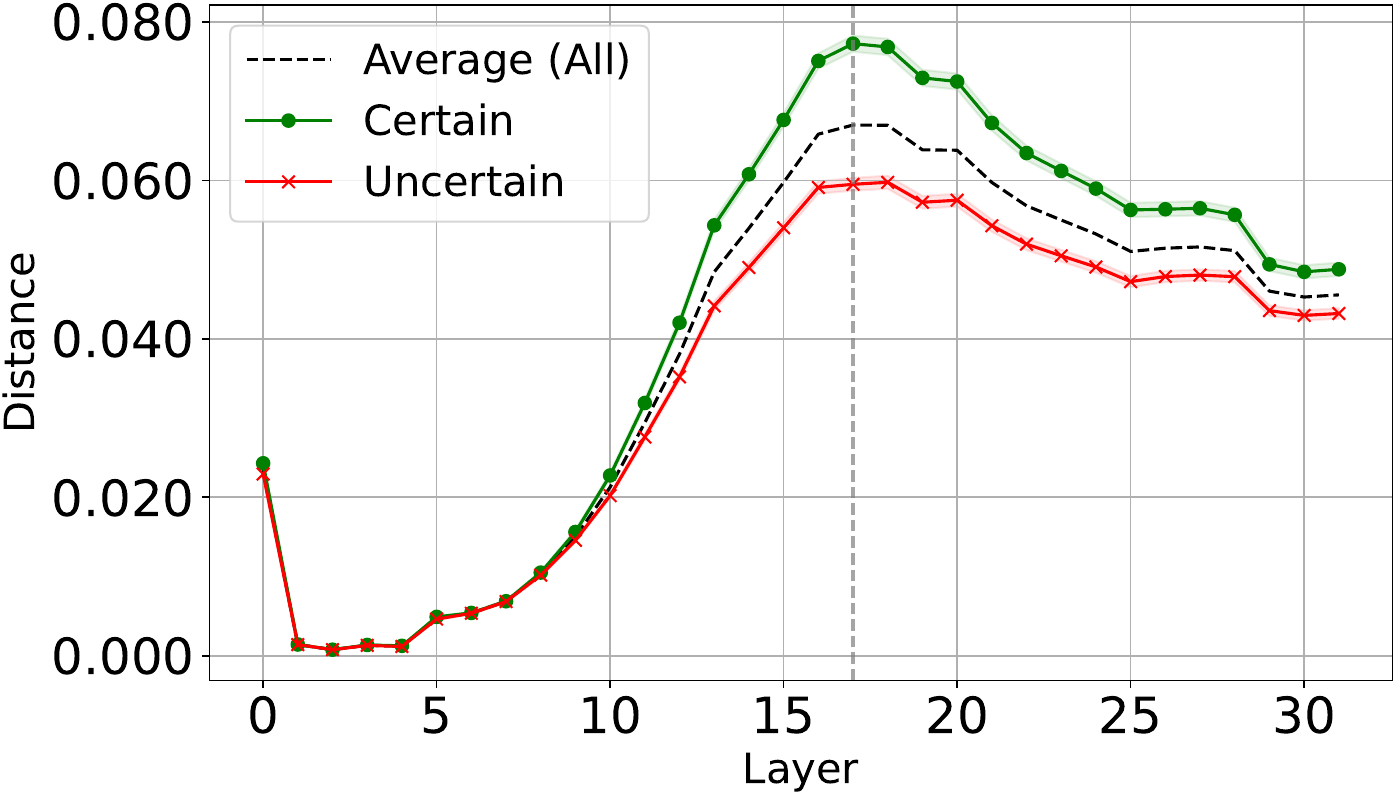}
    \caption{\texttt{idefics-9B}}
\end{subfigure}
\hfill
\begin{subfigure}{0.32\linewidth}
    \centering
    \includegraphics[width=\linewidth]{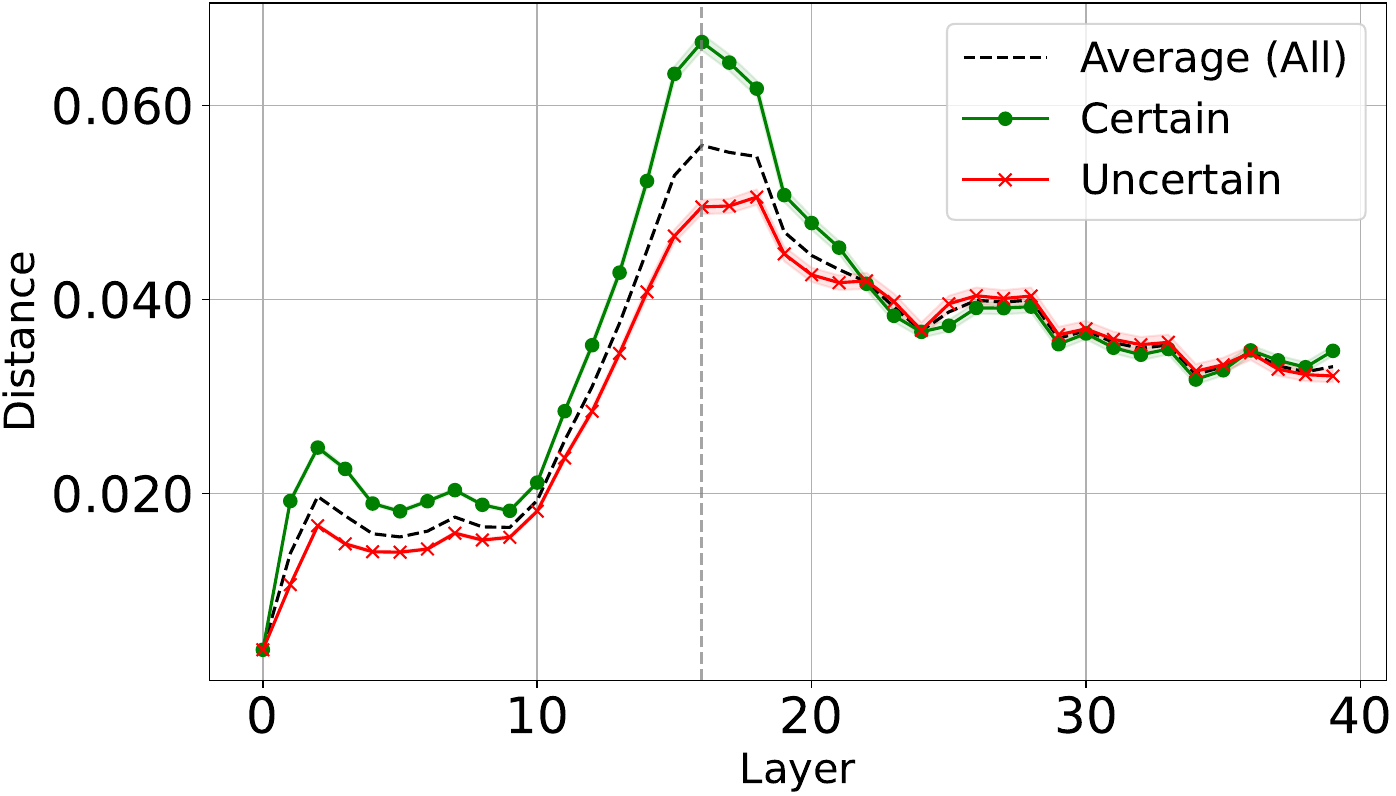}
    \caption{\texttt{llava1.5-13B}}
\end{subfigure}
\hfill
\begin{subfigure}{0.32\linewidth}
    \centering
    \includegraphics[width=\linewidth]{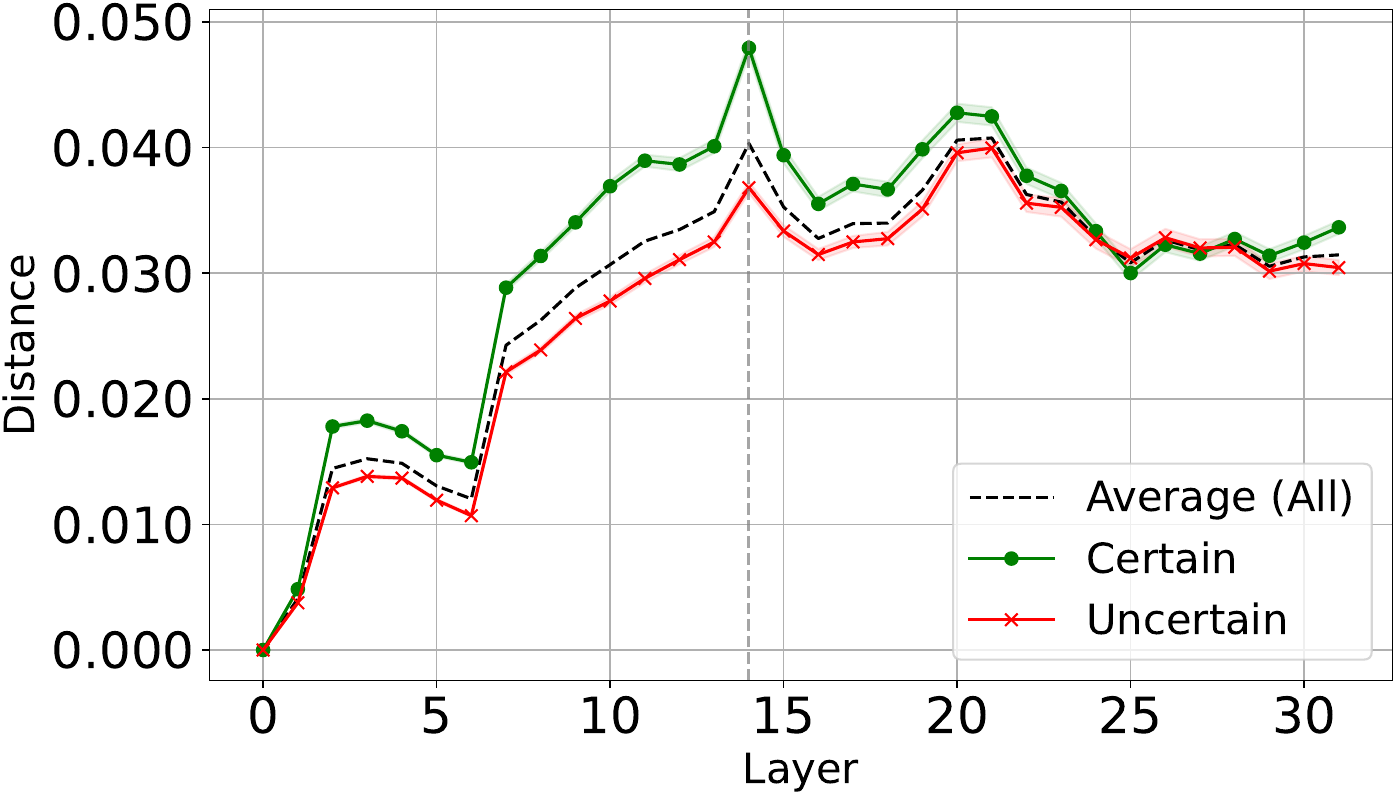}
    \caption{\texttt{llava1.5-7B}}
\end{subfigure}

\vspace{0.6em}

\makebox[\linewidth][c]{%
\begin{subfigure}{0.32\linewidth}
    \centering
    \includegraphics[width=\linewidth]{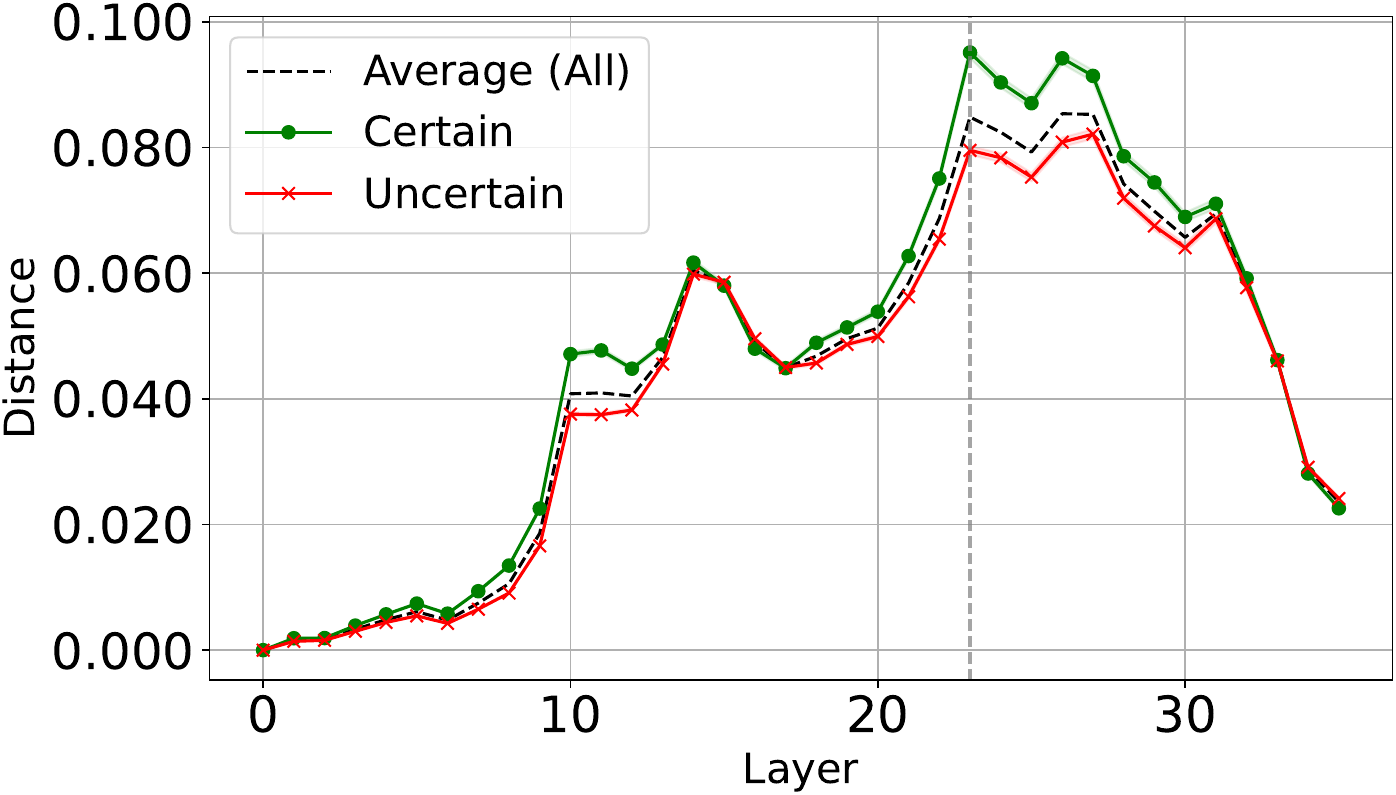}
    \caption{\texttt{Qwen2.5-VL-3B}}
\end{subfigure}
\hspace{0.04\linewidth}
\begin{subfigure}{0.32\linewidth}
    \centering
    \includegraphics[width=\linewidth]{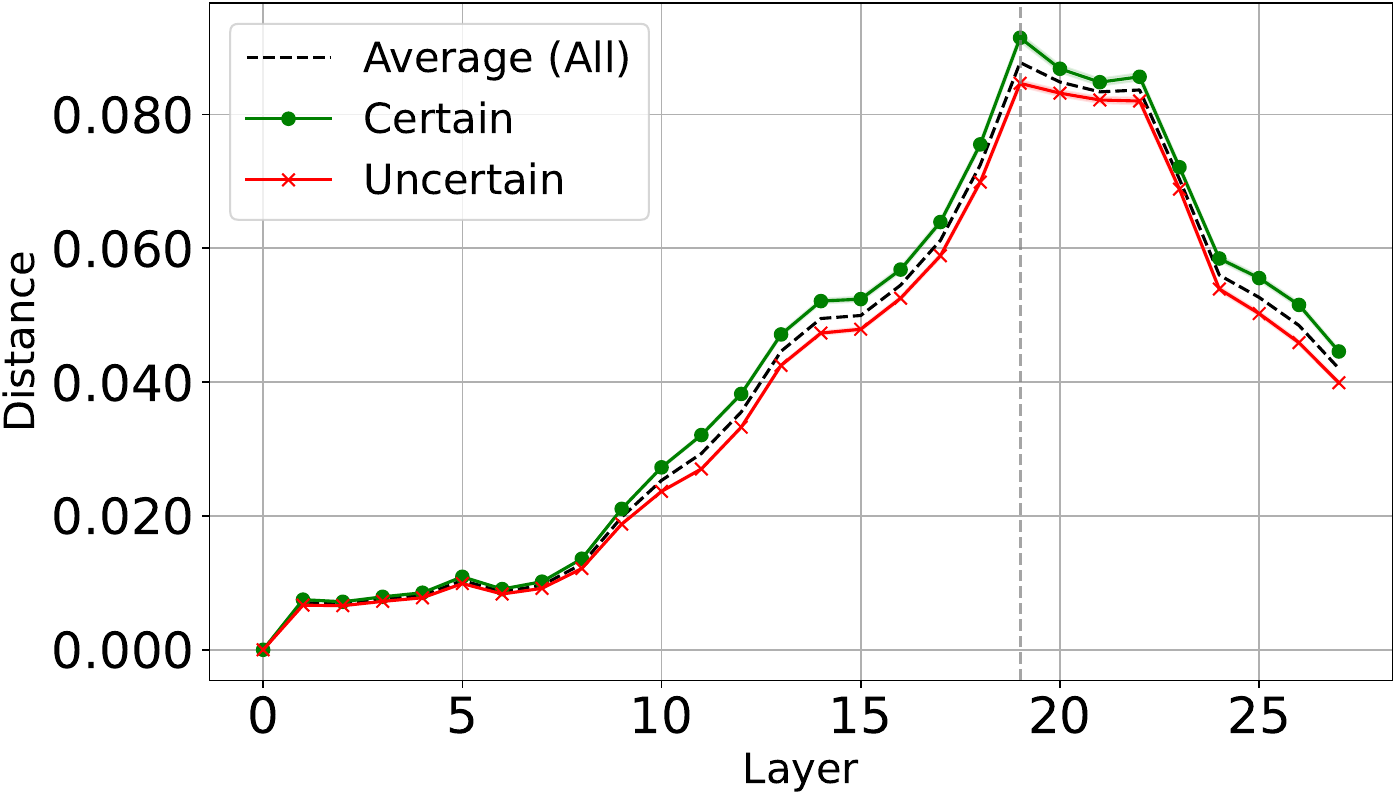}
    \caption{\texttt{Qwen2.5-VL-7B}}
\end{subfigure}
}

\makebox[\linewidth][c]{%
\begin{subfigure}{0.32\linewidth}
    \centering
    \includegraphics[width=\linewidth]{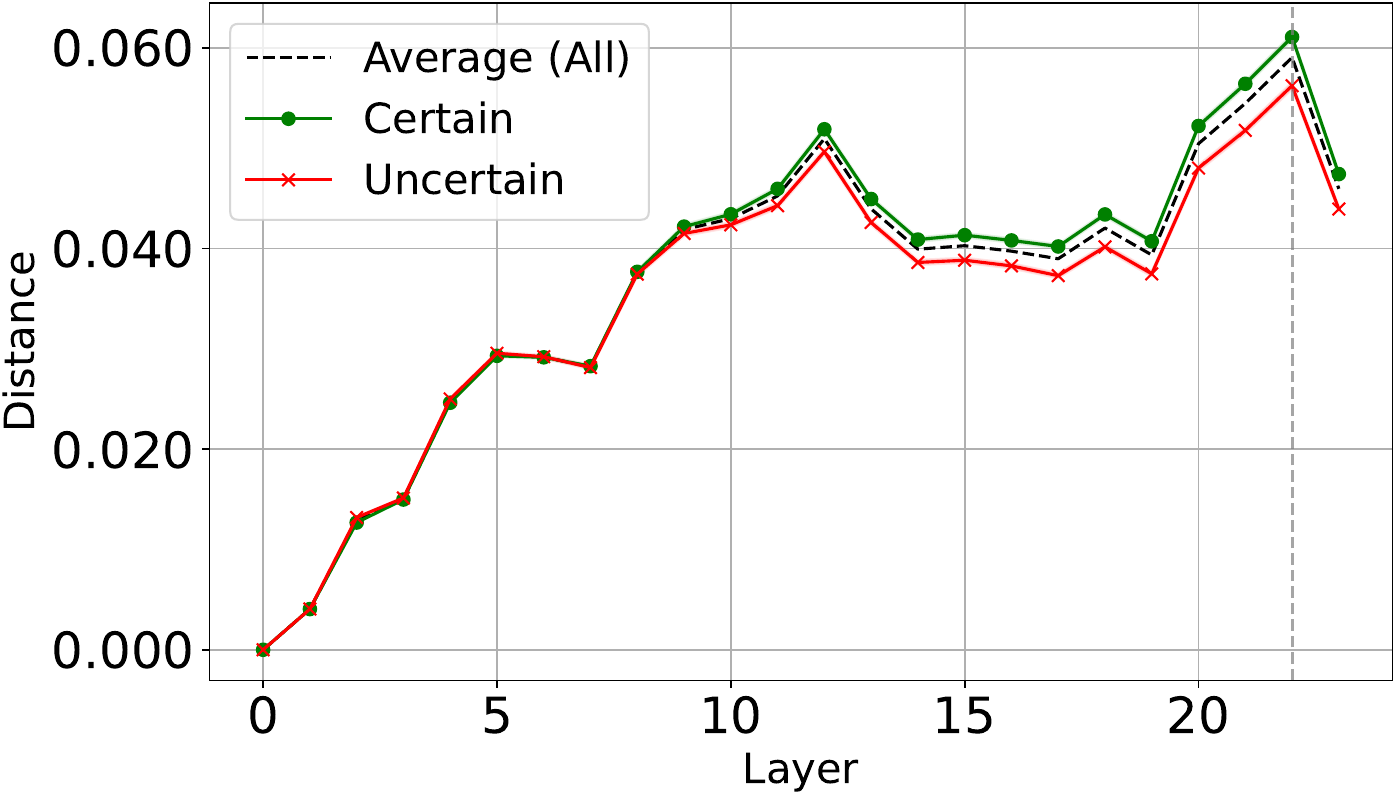}
    \caption{\texttt{Qwen3.5-2B}}
\end{subfigure}
\hspace{0.04\linewidth}
\begin{subfigure}{0.32\linewidth}
    \centering
    \includegraphics[width=\linewidth]{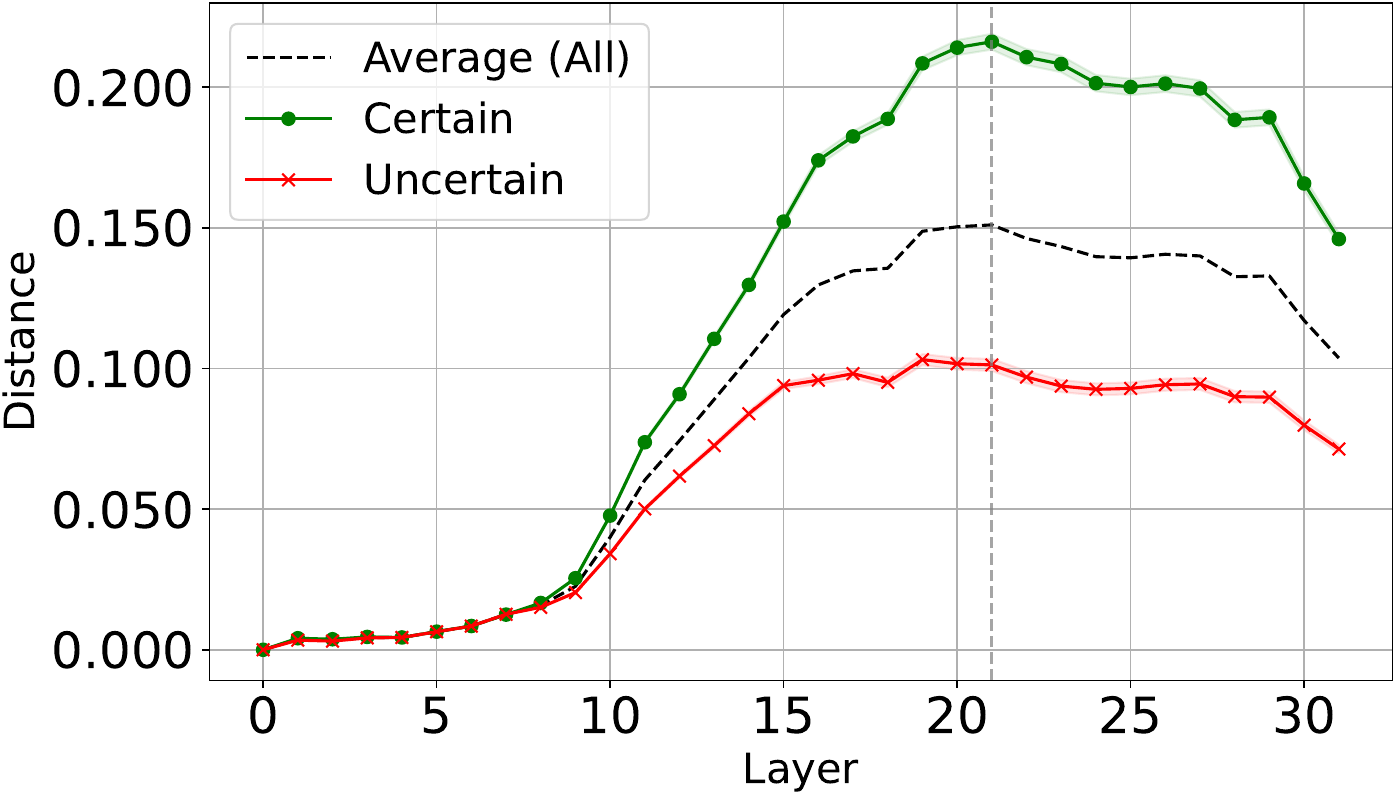}
    \caption{\texttt{idefics2-8B}}
\end{subfigure}
}

\caption{\textbf{Layer-wise distance between hidden states from the original generation pass and the teacher-forced pass with a masked image.} The curves show the average similarity across layers for certain and uncertain samples for different LVLM architectures.}
\label{fig:certain}

\end{figure*}

\paragraph{Analysis wrt prediction certainty.} 
Figure~\ref{fig:certain} shows the evolution of this distance across layers for certain and uncertain predictions. 
Several models are evaluated, including \texttt{idefics2-8B}, \texttt{idefics-9B} \texttt{llava-1.5-13B}, \texttt{llava-1.5-7B}, \texttt{Qwen2.5-VL-3B}, \texttt{Qwen2.5-VL-7B} and \texttt{Qwen3.5-2B}, all evaluated on \textbf{OKVQA} data-set. For all models, we observe that the distance between the two representations is larger for certain predictions than for uncertain ones. 
This means that when the model is confident about its answer, the internal representation changes more when the image is removed.  In other words, confident predictions tend to rely more on visual information. Another observation is that the distance generally increases in deeper layers of the network. This suggests that the influence of visual information becomes stronger in later stages of the Transformer.

\paragraph{Analysis wrt prediction correctness.}

\begin{figure*}[h]
\centering

\begin{subfigure}{0.32
\linewidth}
    \centering
    \includegraphics[width=\linewidth]{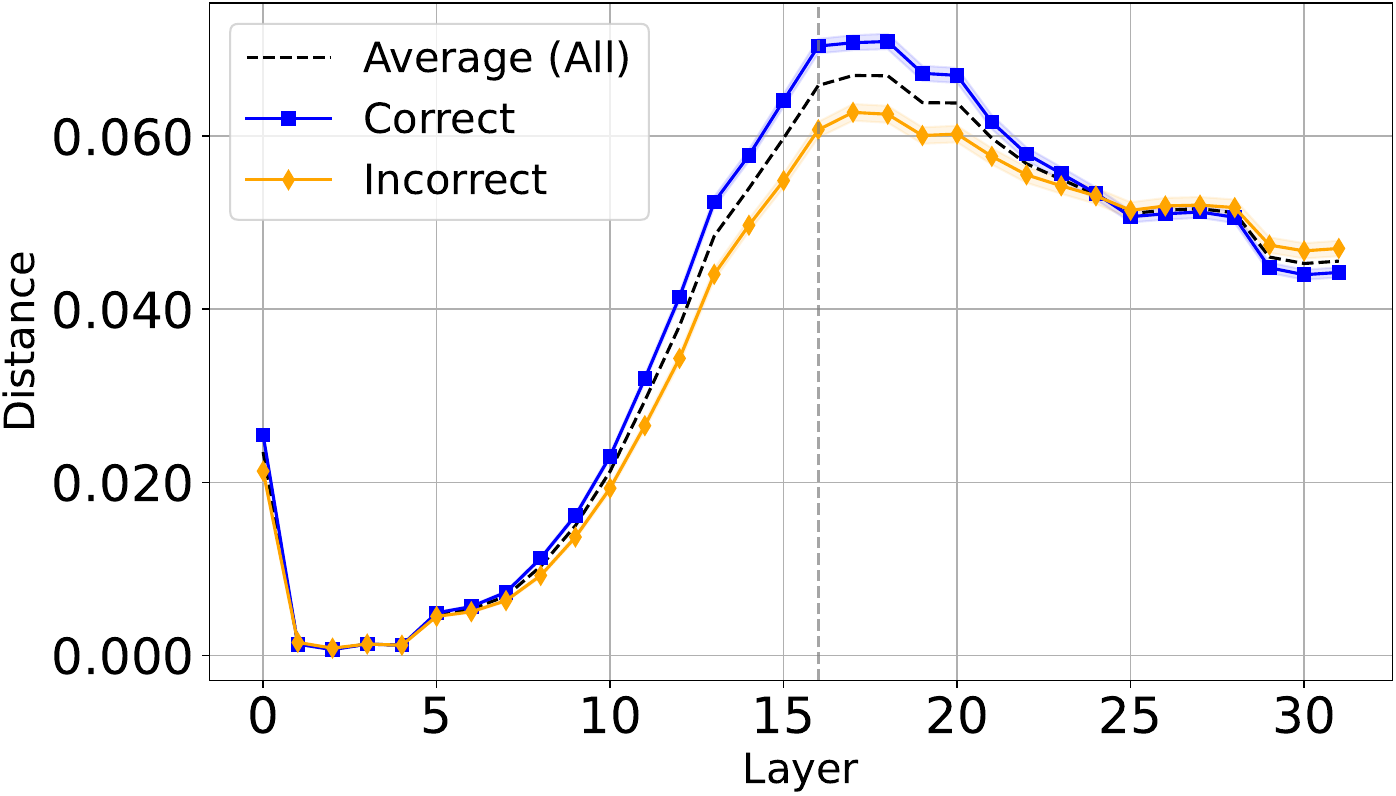}
    \caption{\texttt{idefics-9B}}
\end{subfigure}
\hfill
\begin{subfigure}{0.32\linewidth}
    \centering
    \includegraphics[width=\linewidth]{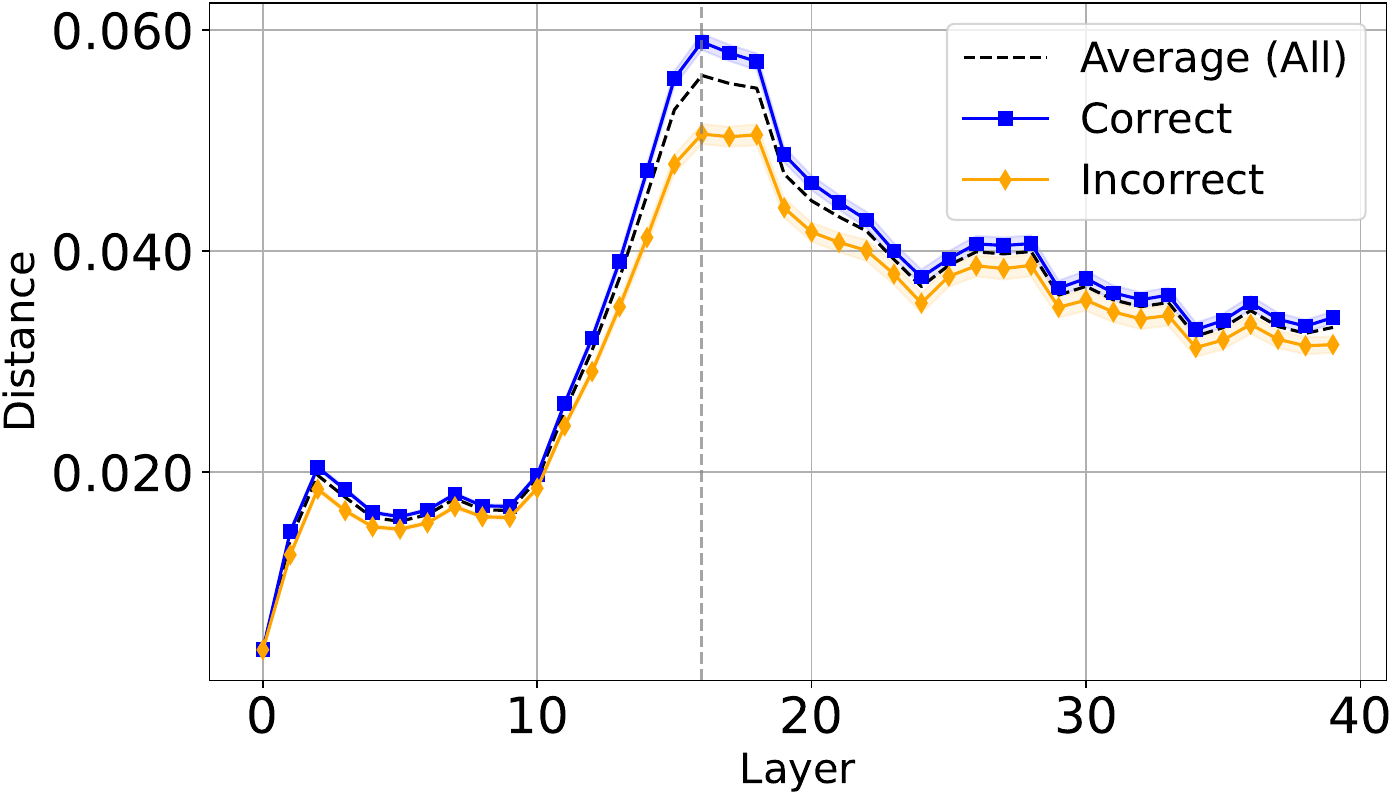}
    \caption{\texttt{llava1.5-13B}}
\end{subfigure}
\hfill
\begin{subfigure}{0.32\linewidth}
    \centering
    \includegraphics[width=\linewidth]{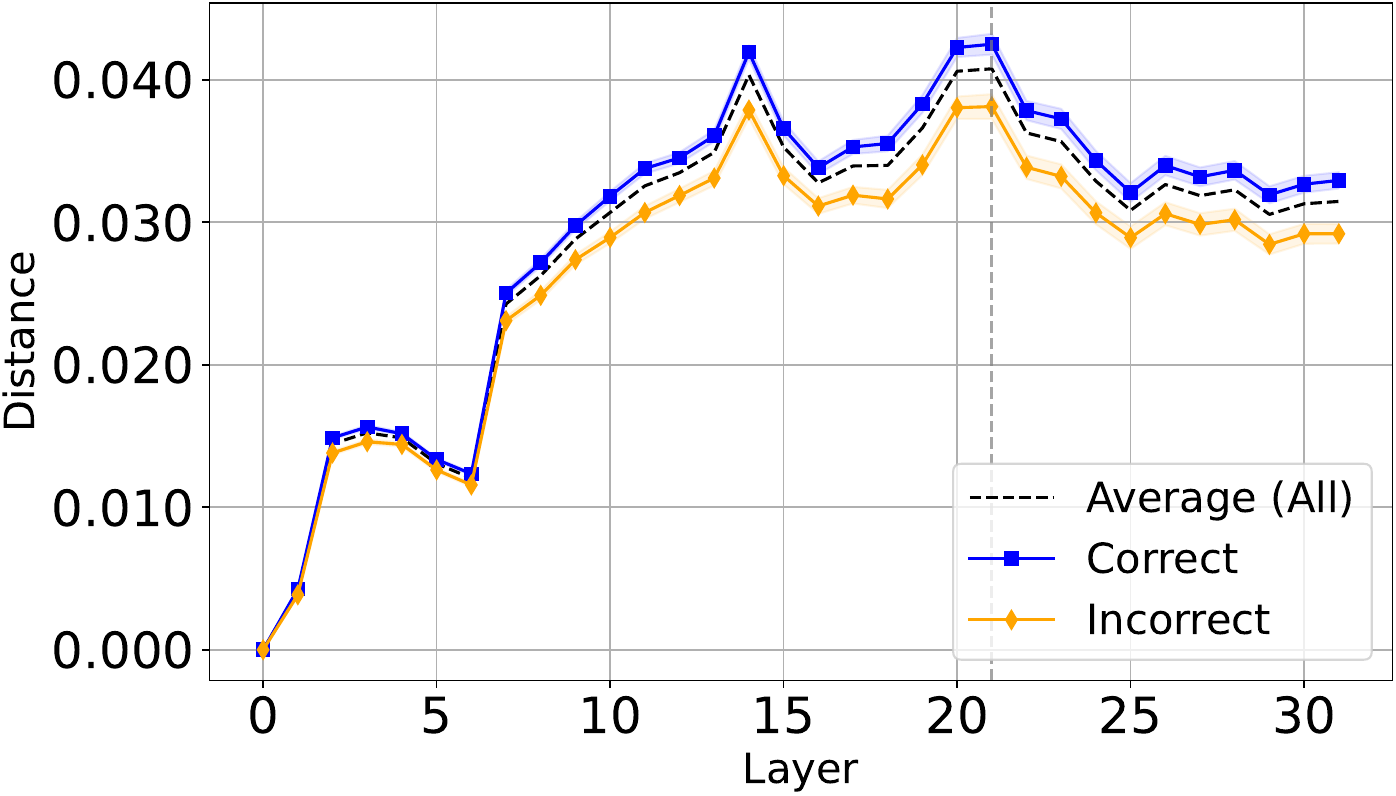}
    \caption{\texttt{llava1.5-7B}}
\end{subfigure}

\vspace{0.6em}

\makebox[\linewidth][c]{%
\begin{subfigure}{0.32\linewidth}
    \centering
    \includegraphics[width=\linewidth]{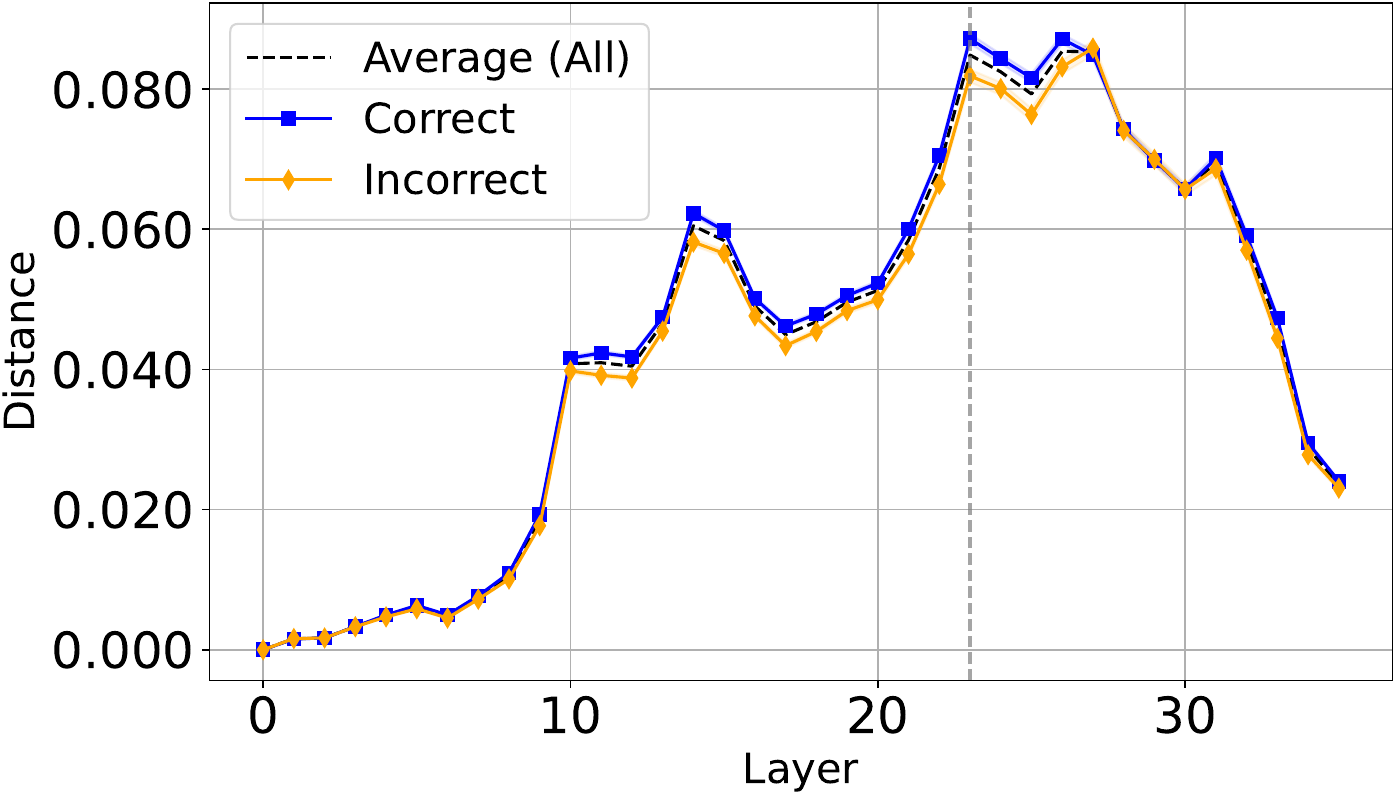}
    \caption{\texttt{Qwen2.5-VL-3B}}
\end{subfigure}
\hspace{0.04\linewidth}
\begin{subfigure}{0.32\linewidth}
    \centering
    \includegraphics[width=\linewidth]{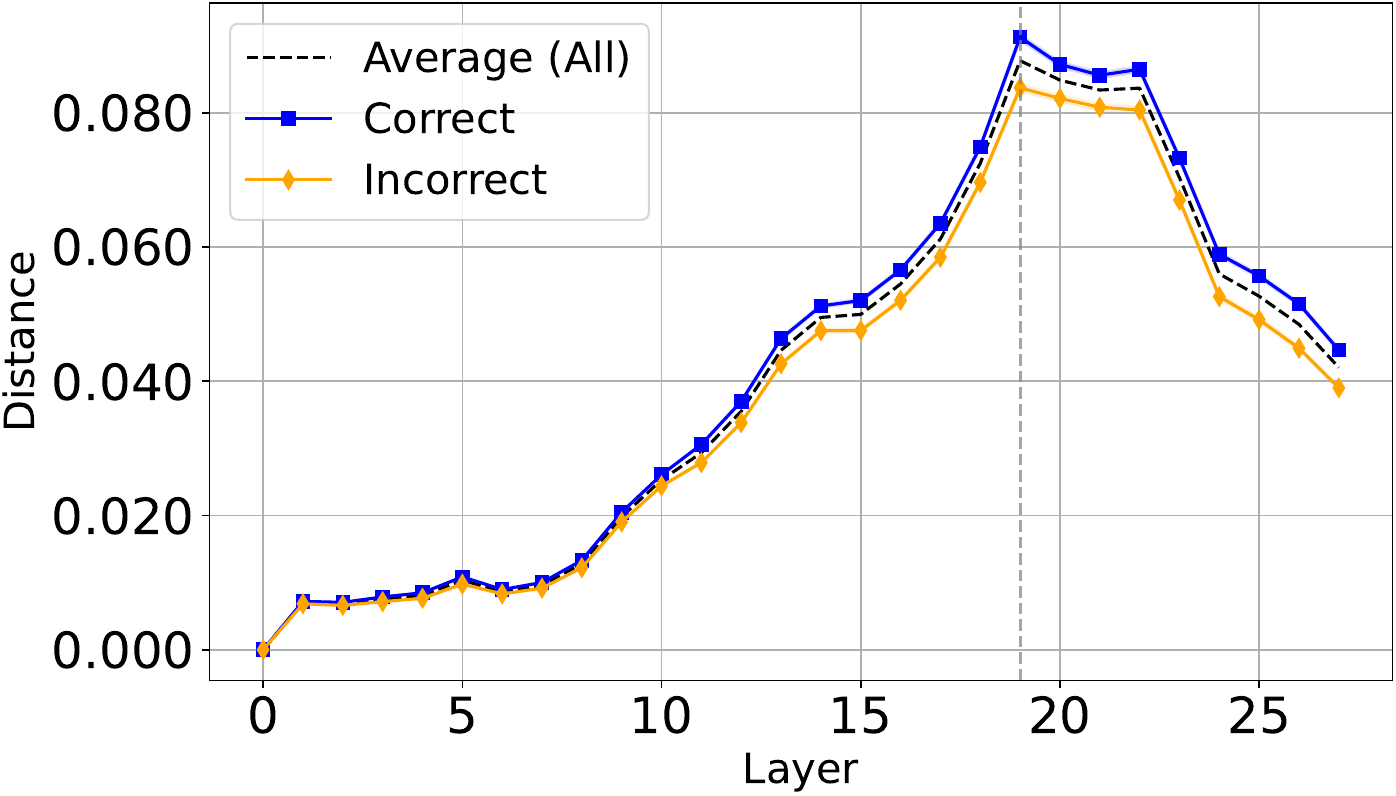}
    \caption{\texttt{Qwen2.5-VL-7B}}
\end{subfigure}
}

\makebox[\linewidth][c]{%
\begin{subfigure}{0.32\linewidth}
    \centering
    \includegraphics[width=\linewidth]{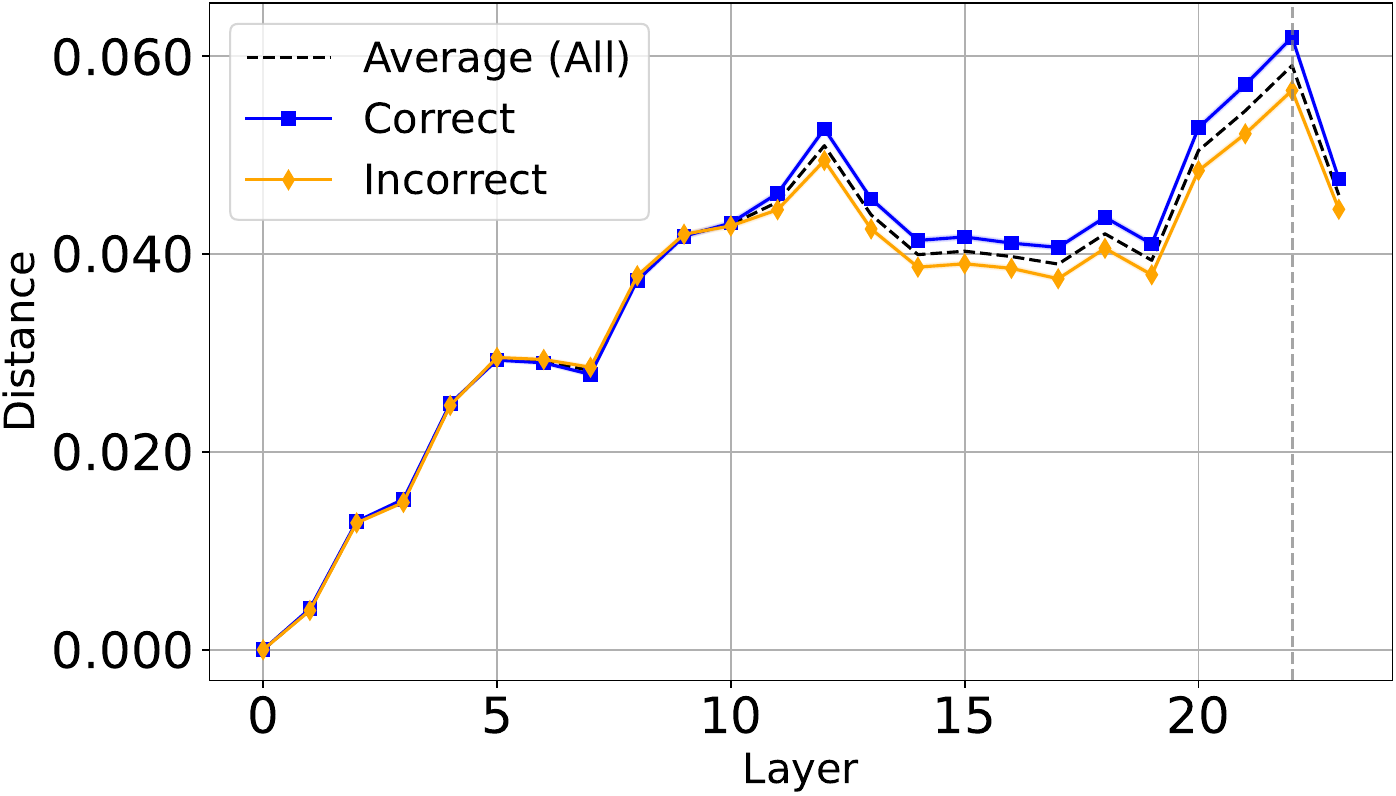}
    \caption{\texttt{Qwen3.5-2B}}
\end{subfigure}
\hspace{0.04\linewidth}
\begin{subfigure}{0.32\linewidth}
    \centering
    \includegraphics[width=\linewidth]{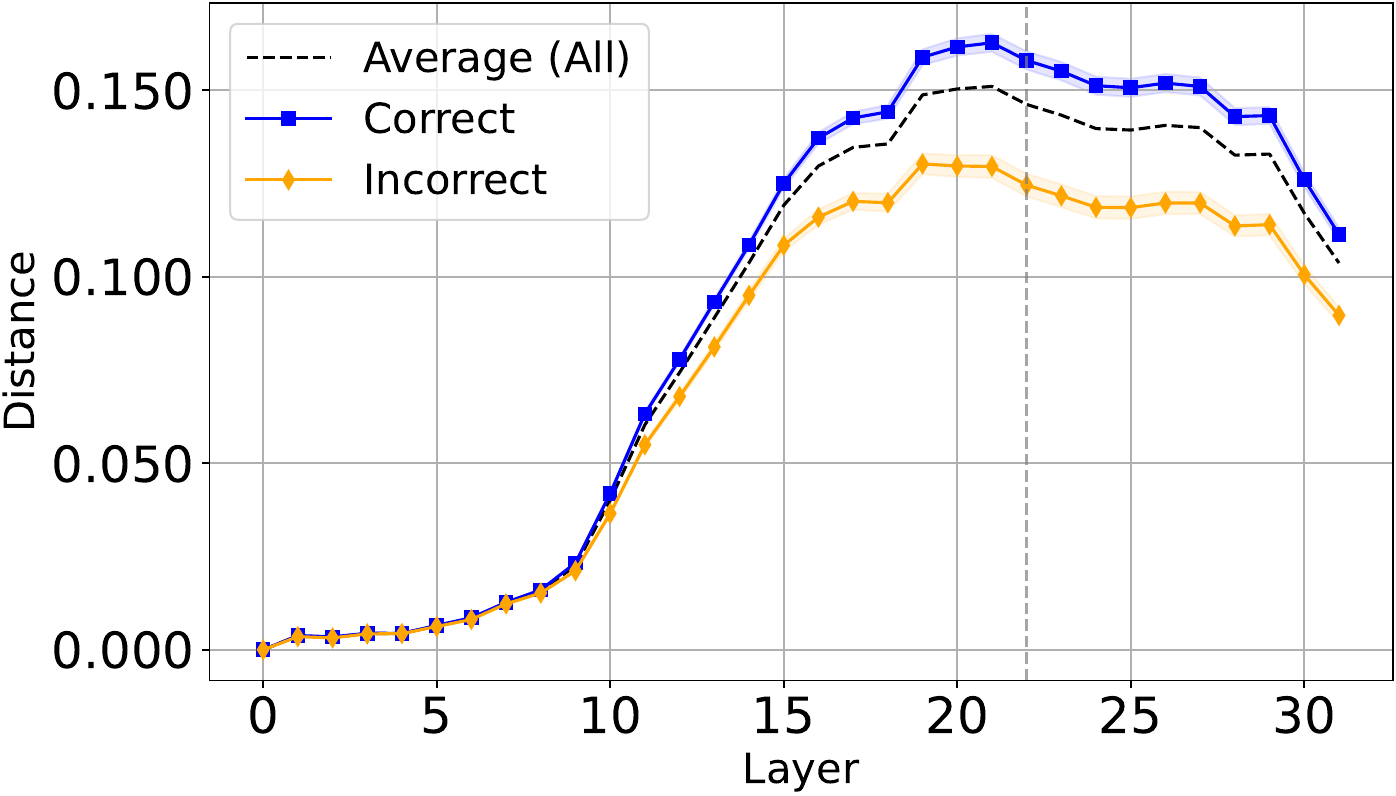}
    \caption{\texttt{idefics2-8B}}
\end{subfigure}
}

\caption{\textbf{Layer-wise distance between hidden states from the original generation pass and the teacher-forced pass with a masked image.} The curves show the average similarity across layers for correct and incorrect samples for different LVLM architectures.}
\label{fig:correct}

\end{figure*}

Figure~\ref{fig:correct} compares the distance between representations for correct and incorrect predictions. For all models, we also observe that the distance between the two representations is larger for correct predictions than for incorrect ones. This indicates that when the model produces the right answer, its internal representation is more strongly affected by removing the image. In other words, correct predictions tend to rely more on visual information. The separation between correct and incorrect predictions is generally slightly smaller than the one observed between certain and uncertain predictions, suggesting that visual reliance is more directly reflected in model confidence than in correctness. As in the certainty analysis, the distance also tends to increase in deeper layers of the network, indicating that visual information has a stronger influence in later Transformer layers.

\paragraph{Additional details on the spider plot.}
\label{app_spider_plot}

Figure~\ref{fig:spider} provides a compact aggregation of the layer-wise analyses reported in Figures~\ref{fig:certain} and~\ref{fig:correct}. To summarize each comparison in the spider plot, we select the layer at which the absolute gap between the two group-wise mean curves is maximal. This selected layer corresponds to the dashed vertical line in Figures~\ref{fig:correct} and~\ref{fig:certain}. We then extract the two group values at this layer, jointly normalize them within the corresponding comparison, and use them as the radial values for the associated model in Figure~\ref{fig:spider}. Thus, each spoke in Figure~\ref{fig:spider} summarizes, for one model, the strongest layer-wise separation between the two groups, with larger polygon separation indicating that the image-induced representation shift is more discriminative of correctness or certainty.

\end{document}